\pdfoutput=1

\documentclass[11pt]{article}

\usepackage{acl}

\usepackage{lipsum}
\usepackage{microtype,textcase}
\usepackage{mathtools,xparse}
\usepackage{amsthm, physics}
\usepackage{cleveref}
\usepackage{amsmath}
\usepackage{algorithm,enumitem}
\usepackage{algpseudocode}
\usepackage{algorithmicx}
\usepackage{multirow}
\usepackage{graphicx}
\usepackage{booktabs}
\usepackage{pifont}
\usepackage{xcolor}
\usepackage{blindtext}
\usepackage{colortbl}
\usepackage{tabulary}
\usepackage{etoolbox}
\usepackage{hyperref}
\usepackage{balance}
\usepackage{times}
\usepackage{latexsym}
\usepackage{subfig}
\usepackage{tabularx}
\usepackage{pbox}
\usepackage{float}
\usepackage{setspace}
\usepackage{diagbox}
\usepackage{caption}
\usepackage{bbding}
\usepackage{enumitem}
\usepackage{mathtools}
\usepackage{color, colortbl}
\definecolor{grey}{rgb}{0.898,0.898,0.898}
\usepackage{amssymb}
\newcommand{\proposed}{\textsc{Diva}}
\usepackage{array}
\usepackage{url}
\usepackage{tikz}
\usepackage{bbm}
\usepackage{dsfont}
\usepackage{wrapfig}
\usepackage{titletoc}
\usepackage[T1]{fontenc}
\usepackage{ulem}
\definecolor{gainsboro}{RGB}{233,233,233}
\usepackage[utf8]{inputenc}
\usepackage{microtype}
\usepackage{inconsolata}
\usepackage{paralist}

\algnewcommand\algorithmicinput{\textbf{Input:}}
\algnewcommand\Input{\item[\algorithmicinput]}
\algnewcommand\algorithmicoutput{\textbf{Output:}}
\algnewcommand\Output{\item[\algorithmicoutput]}

\title{Diversify-verify-adapt: Efficient and Robust Retrieval-Augmented Ambiguous Question Answering}

\author{
\textbf{Yeonjun In}\textsuperscript{1}\thanks{Work done during internship at Adobe Research.}, \ 
\textbf{Sungchul Kim}\textsuperscript{2}\thanks{Corresponding author.}, \ 
\textbf{Ryan A. Rossi}\textsuperscript{2}, \ 
\textbf{Md Mehrab Tanjim}\textsuperscript{2}, \\
\textbf{Tong Yu}\textsuperscript{2},\
\textbf{Ritwik Sinha}\textsuperscript{2},\
\textbf{Chanyoung Park}\textsuperscript{1}\\
\textsuperscript{1}KAIST \qquad
\textsuperscript{2}Adobe Research \\
\texttt{\{yeonjun.in, cy.park\}@kaist.ac.kr} \\
\texttt{\{sukim, ryrossi, tanjim, tyu, risinha\}@adobe.com}}

\begin{document}
\maketitle
\begin{abstract}
The retrieval augmented generation (RAG) framework addresses an ambiguity in user queries in QA systems by retrieving passages that cover all plausible interpretations and generating comprehensive responses based on the passages. However, our preliminary studies reveal that a single retrieval process often suffers from low-quality results, as the retrieved passages frequently fail to capture all plausible interpretations. Although the iterative RAG approach has been proposed to address this problem, it comes at the cost of significantly reduced efficiency.
To address these issues, we propose the \underline{\textbf{di}}versify-\underline{\textbf{v}}erify-\underline{\textbf{a}}dapt (\proposed) framework. \proposed~first \textbf{diversifies} the retrieved passages to encompass diverse interpretations. Subsequently, \proposed~\textbf{verifies} the quality of the passages and \textbf{adapts} the most suitable approach tailored to their quality.
This approach improves the QA systems' accuracy and robustness by handling low quality retrieval issue in ambiguous questions, while enhancing efficiency.

\end{abstract}

\section{Introduction}
\label{sec:introduction}

Open-domain question answering (QA) systems aim to provide factual responses across diverse topics. However, ambiguity in user queries is common, with over 50\% of Google search queries falling into this category \cite{min2020ambigqa}. Ambiguous questions challenge QA systems to determine user intent, making it essential for them to deliver answers covering all possible interpretations.



Addressing ambiguous questions is crucial in real-world applications, yet remains underexplored compared to unambiguous questions \cite{joshi-etal-2017-triviaqa, kwiatkowski-etal-2019-natural}. This work aims to fill this gap by tackling the complexities of ambiguous QA.

\begin{figure}
    \centering
    \includegraphics[width=1.\columnwidth]{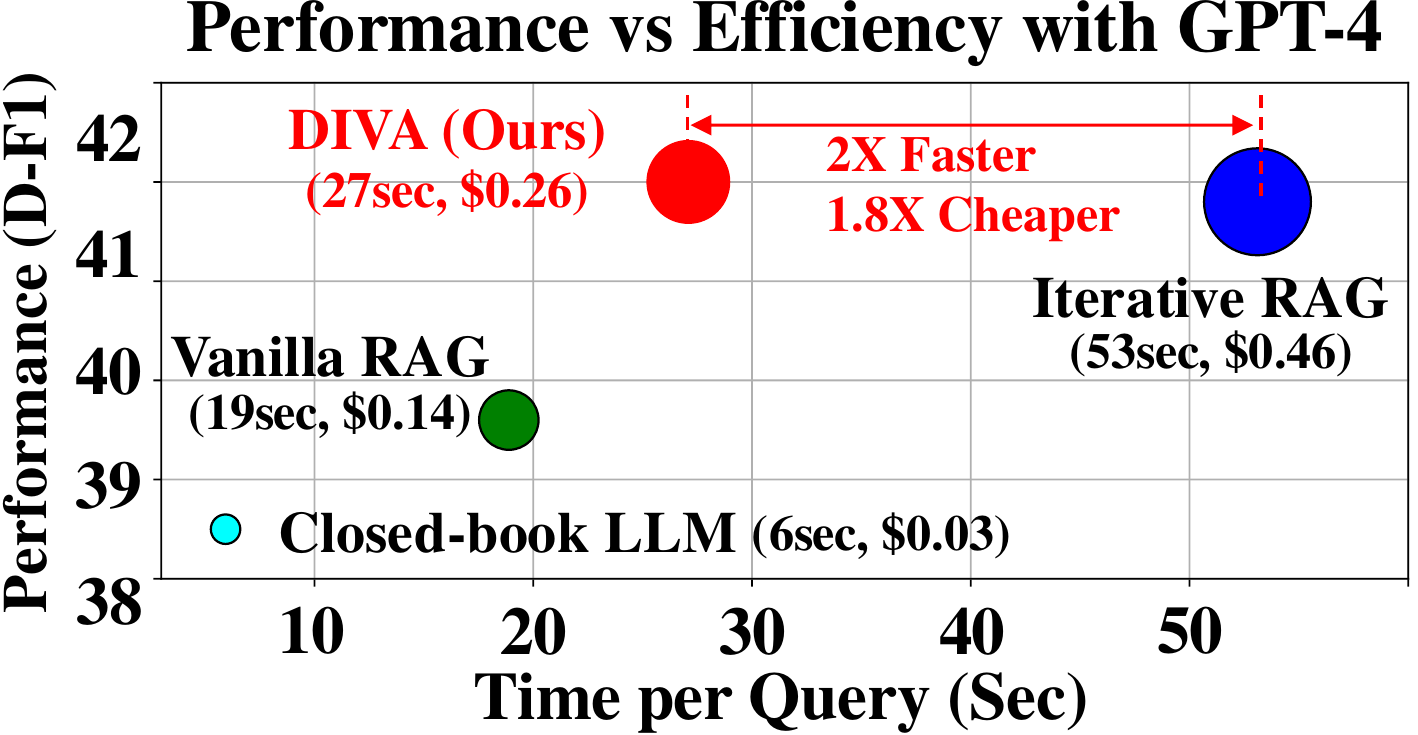}
    \vspace{-2ex}
    \caption{Trade-off between performance and efficiency under GPT-4 backbone on ASQA. Notably, \proposed~achieves better performance to the iterative RAG \cite{kim2023tree},
    while significantly more efficient (that is, 2x faster and 1.8x cheaper). The size of the circle indicates the cost per query (\$). Closed-book LLM indicates the traditional few-shot prompting method used in \citet{brown2020language}
    }
    \vspace{-4ex}
    \label{fig:efficiency} 
\end{figure}

\begin{figure*}
    \centering
    \includegraphics[width=1.\textwidth]{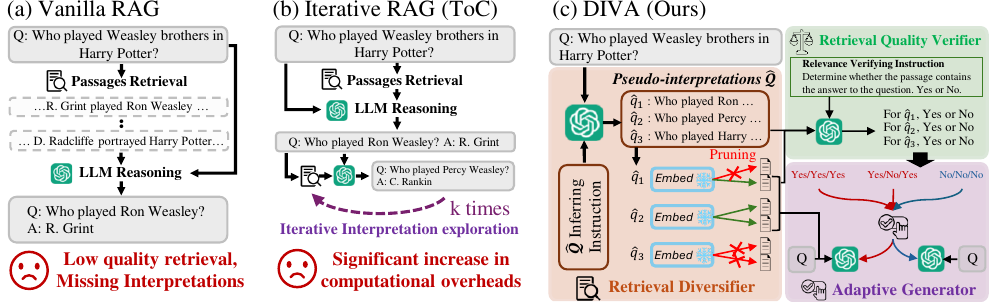}
    \caption{A conceptual comparison of RAG approaches to ambiguous QA. (a) Vanilla RAG retrieves passages and generates answers in a single pass, but it may not collect enough information for diverse interpretations (i.e., low-quality retrieval), compromising factual accuracy. (b) 
    Iterative RAG retrieves passages and generates answers in a loop, using previous interpretations to enhance each subsequent iteration's retrieval for exploring missing interpretations. Although effective, it is inefficient due to the repeated use of LLMs and retrievers.
    (c) \proposed~retrieves passages covering diverse interpretations without relying on the iterative process and selects the most suitable knowledge for response generation by verifying retrieval quality.
    }
    
    \label{fig:overview} 
    \vspace{-3ex}
\end{figure*}

Retrieval-augmented generation (RAG) framework has made significant progress in open-domain QA tasks \cite{izacard-grave-2021-leveraging,lazaridou2022internet,shi2023replug, ram2023context} and also proven to be an effective solution for addressing ambiguous questions \cite{min2020ambigqa, min2021joint, kim2023tree, iterativeprompting}. Specifically, these approaches first retrieve passages on the given question and prompt the LLM to extract plausible interpretations and answers relying on the passages (c.f. Fig~\ref{fig:overview}(a)).

Despite the success of the RAG framework on the ambiguous QA task, we should rethink:
Is a single retrieval process sufficient to retrieve passages encompassing all plausible interpretations? To answer this question, we conduct preliminary experiments (c.f. Sec~\ref{sec:preliminary}) about the quality of the retrieved passages used in the  RAG framework. We observe that the passages obtained from the single retrieval process often pose a low quality issue with respect to addressing ambiguous questions. In other words, the retrieved passages often partially or completely failed to cover all plausible interpretations, leading to significant performance degradation in terms of factual accuracy. 

To address this issue, the iterative RAG approach, ToC \cite{kim2023tree}, has been introduced (c.f. Fig~\ref{fig:overview}(b)) to further explore other interpretations that can not be covered by the single retrieval process. Specifically, 
to further explore missing interpretations, the interpretations extracted in the previous iteration are utilized as queries to retrieve new passages, additional interpretations are then extracted. This exploration process is repeated in multiple times, leading to encompassing more diverse interpretations and corresponding answers. However, we argue that this effectiveness comes with a significant increase in computational overheads due to the iterative passage retrieval and LLM reasoning. In our experiments, this method requires an average of 5.5 exploration steps per query. As shown in Figure~\ref{fig:efficiency}, Iterative RAG (i.e., ToC) significantly outperforms the vanilla RAG approach in terms of factual accuracy but at the cost of greatly reduced efficiency, with notable increases in both inference time and API call costs.

To this end, we introduce an efficient and robust RAG framework for ambiguous QA, referred to as \underline{\textbf{di}}versify-\underline{\textbf{v}}erify-\underline{\textbf{a}}dapt (\proposed). \proposed~comprises two key components efficiently addressing the low quality retrieval issue: \textbf{1)} Retrieval Diversification (RD) and \textbf{2)} Adaptive Generation (AG). The key idea of RD is to infer \textit{pseudo-interpretations} of a question, using them to retrieve a set of passages that broadly cover these interpretations, thus enhancing retrieval quality without any iterative interpretation exploration process. To further enhance the robustness of this framework, we propose an adaptive generation (AG) method. The key idea of AG is to carefully {verify} the overall quality of the passages retrieved from RD before indiscriminately incorporating them. More specifically, we define a new criterion of quality levels tailored to ambiguous questions: $\{\texttt{Useful}, \texttt{PartialUseful}, \texttt{Useless}\}$. Subsequently, AG adapts the most suitable approach between relying on the retrieved passages and LLM's internal knowledge, each of which is tailored to the specific quality level of the passages.


Experiments demonstrate that the proposed RD method efficiently diversifies the retrieval process to obtain passages covering diverse interpretations, thereby enhancing both QA and retrieval accuracy. Additionally, the proposed AG method successfully discriminate low quality passages, leading to the improvement of the QA performance. Consequently, \proposed~outperforms existing baselines on the ASQA \cite{stelmakh2022asqa} and SituatedQA \cite{zhang-choi-2021-situatedqa} across various    LLM backbones in a few-shot setup, achieving superior accuracy and efficiency. The key contributions of this work are as follows: 



\begin{compactitem}
    
    \item To the best of our knowledge, this paper is the first attempt to investigate the practical limitations of the existing RAG frameworks {when applied to ambiguous QA task}: low quality retrieval and inefficiency.
    \item We propose \proposed, an efficient and robust RAG framework that efficiently retrieves diverse passages, verifies their quality, and adapts the most suitable approach tailored to each retrieval quality.
    \item \proposed~consistently outperforms state-of-the-art RAG approaches in ambiguous QA task, while significantly more efficient (nearly 1.5 - 3 times faster response generation). 
    
\end{compactitem}

\section{Preliminary Experiments}
\label{sec:preliminary}

We investigate the quality of retrieved passages and their impact on the performance of the RAG framework (as in Fig~\ref{fig:overview}(a)) in ambiguous QA task.

\noindent \textbf{Experimental Details.} \@
We utilize the most recent ambiguous QA dataset, ASQA \cite{stelmakh2022asqa}. We classify the quality of retrieved passages into three labels: \textbf{1)} Fully Cover, \textbf{2)} Partially Cover, and \textbf{3)} Not Cover. Fully Cover indicates that the retrieved passages encompass all plausible interpretations, Not Cover does that the retrieved passages do not contain any of them, and otherwise Partially Cover.  We obtain these labels for each question by computing a string exact match between a set of retrieved passages and all plausible answers provided in ASQA as ground-truth answers. For implementation details of retrieving passages, see Appendix~\ref{sec:ap:retrieval-process}.

\begin{figure}
    \centering
    \includegraphics[width=\columnwidth]{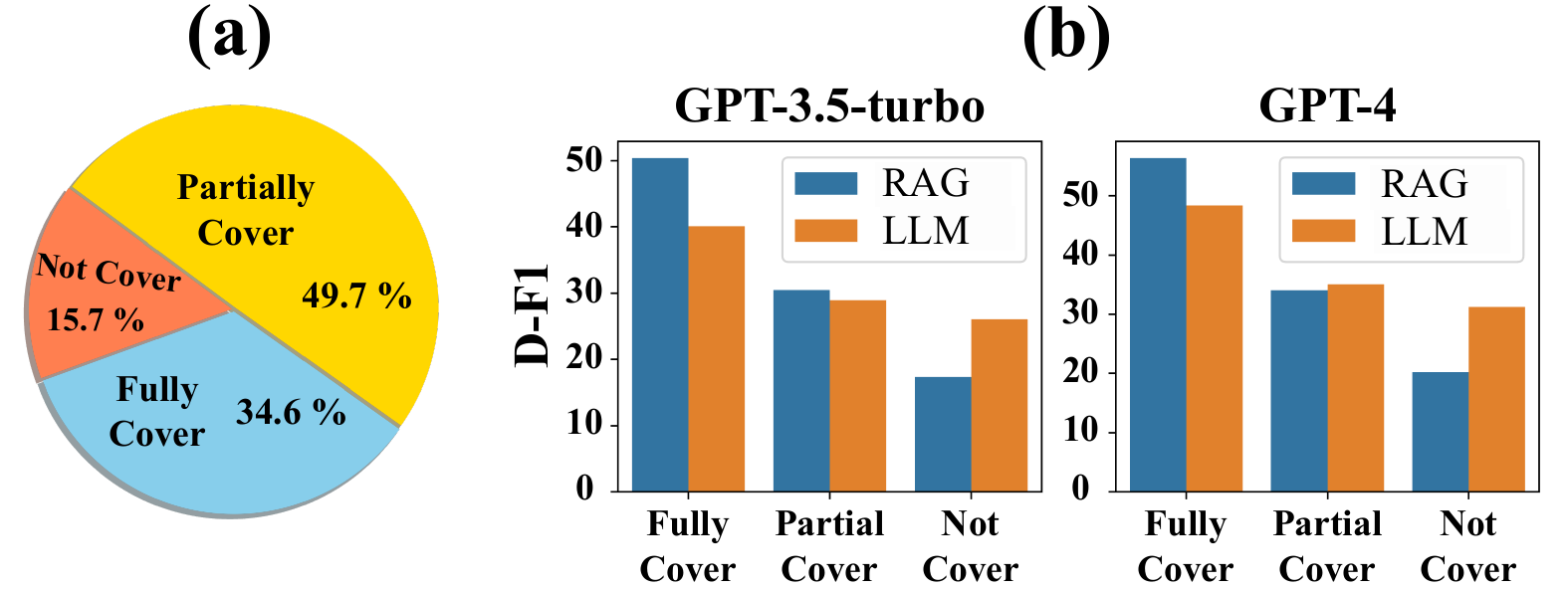}
    \vspace{-4ex}
    \caption{Preliminary results on ASQA. \textbf{(a)} Portion of each quality label of retrieved passages. 
    \textbf{(b)} Performance comparison upon the quality label. 
    }
    
    \label{fig:preliminary_studies} 
    \vspace{-2ex}
\end{figure}

\noindent \textbf{Results.} \@
In Fig~\ref{fig:preliminary_studies}(a), we observe that for only 34.6\% of questions (i.e., Fully Cover) the retriever successfully retrieves passages that cover all plausible interpretations. Additionally, for 15.7\% of questions (i.e., Not Cover) the retriever fails to retrieve any relevant passages. 
More critically, as shown in Fig~\ref{fig:preliminary_studies}(b), the performance of the RAG framework (i.e., RAG in the figure)
significantly deteriorates in terms of the factual accuracy (i.e., D-F1) when the retrieved passages pose a low quality issue (i.e., Partial Cover and Not Cover), indicating that it is highly susceptible to noise and irrelevant information in the ambiguous QA.

This observation raises a follow-up question: How can we handle cases where the retrieved passages do not fully cover the plausible answers? To address this issue, we conducted another experiment that compares the effectiveness of LLM's internal knowledge and provided passages for different cases, respectively. We observe that when the retrieved passages do not contain any of the plausible interpretations (i.e., Not Cover), the closed-book LLM (i.e., LLM in the figure) significantly outperforms the RAG framework. This suggests that QA performance benefits more from relying on the LLMs' internal knowledge rather than on external passages containing entirely irrelevant information.

In short, while the quality of retrieval is crucial for the performance of the RAG framework in ambiguous QA, existing works have largely overlooked this critical issue, which notably diminishes their practical applicability.

\section{Proposed Method: \proposed}
\label{sec:method}

Based on the findings, we propose an efficient and robust RAG framework for ambiguous QA, \textbf{\underline{di}}versify-\textbf{\underline{v}}erify-\textbf{\underline{a}}dapt (\proposed).
This framework comprises two key components: Retrieval Diversification \textbf{(Sec~\ref{sec:retreival-diversification})} and Adaptive Generation \textbf{(Sec~\ref{sec:retrieval-verification})}. The retrieval diversification method aims to efficiently \textbf{diversify} the retrieved passages to encompass diverse interpretations. Subsequently, the adaptive generation method aims to \textbf{verify} the quality of the passages and \textbf{adapt} the most suitable approach tailored to their quality. Fig~\ref{fig:overview}(c) and Algorithm~\ref{alg:algo1} show the overview and inference algorithm of~\proposed, respectively.

\subsection{Problem Formulation}
\label{sec:problem_formuation}

Given an ambiguous question $q_i$, the goal of the proposed RAG framework is to generate a comprehensive response $r_i$ that encompasses all plausible answers $\mathcal{A}_i= \{ a_{i,1},...,a_{i,M} \}$ of the interpretations $\mathcal{Q}_i=\{ q_{i,1},...,q_{i,M} \}$ based on the retrieved passages $\mathcal{P}_i = \{ p_{i,1}, ..., p_{i, K} \}$, where $M$ and $K$ indicate the number of plausible answers and passages, respectively. Specifically, given the $\mathcal{P}_i$ ideally contains all $\mathcal{Q}_i$ and $\mathcal{A}_i$, an LLM is first prompted with the question and the relevant passages to extract all plausible interpretations and their corresponding answers, formally represented as follows:

\vspace{-3ex}
\begin{equation}
\mathcal{Q}_i, \mathcal{A}_i \leftarrow \texttt{LLM}(q_i, \mathcal{P}_i, I_{\text{e}}),
\label{eq:disambiguation}
\end{equation}
\vspace{-3ex}

\noindent where $I_\text{e}$ is a text prompt for extracting $\mathcal{Q}_i$ and $\mathcal{A}_i$ from the $\mathcal{P}_i$. Subsequently, based on the $\mathcal{Q}_i$ and $\mathcal{A}_i$, the LLM is prompted to consolidate them with $q_i$ and $\mathcal{P}_i$ to generate a response $r_i$, formally represented as follows:

\vspace{-2ex}
\begin{equation}
r_i \leftarrow \texttt{LLM}(\mathcal{Q}_i, \mathcal{A}_i, \mathcal{P}_i, q_i, I_\text{g}).
\label{eq:question-answering}
\end{equation}
\vspace{-3ex}

\noindent For the prompts $I_\text{e}$ and $I_\text{g}$, we start with that of \citet{kim2023tree} and modify it for our setup (see Table~\ref{tab:ap:prompt_extract_interpretations_from_passage} and Table~\ref{tab:ap:prompt_rag_response_generation} in Appendix~\ref{sec:ap:prompt}).


\subsection{Retrieval Diversification (RD)}
\label{sec:retreival-diversification}

In this section, we propose a novel retrieval diversification (RD) method aiming to efficiently identify passages $\mathcal{P}_i$ encompassing all plausible answers $\mathcal{A}_i$ of the interpretations $\mathcal{Q}_i$. The key idea of RD is to infer \textit{pseudo-interpretations}\footnote{We define \textit{pseudo-interpretations} as approximate interpretations closely resembling the actual interpretations.} of a question, using them to retrieve a set of passages that maximally cover these interpretations. This approach guarantees the retrieved passages encompass diverse interpretations, without any iterative interpretation exploration process of \citet{kim2023tree}, leading to the generated response $r_i$ covering all $\mathcal{A}_i$.




\begin{figure}
    \centering
    \includegraphics[width=0.9\columnwidth]{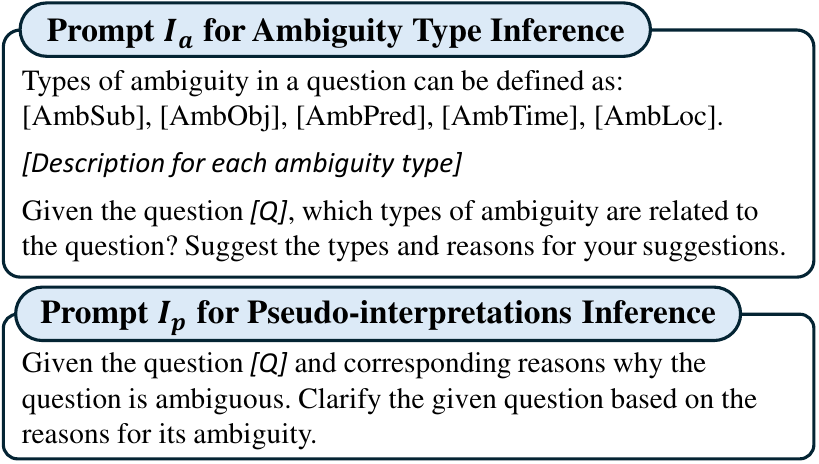}
    \vspace{-1ex}
    \caption{Conceptual example of prompts for pseudo-interpretations inference.
    }
    
    \label{fig:conceptual_prompt} 
    \vspace{-2ex}
\end{figure}

\noindent \textbf{Inferring Pseudo-Interpretations.} \@ 
To infer \textit{pseudo-interpretations} $\hat{\mathcal{Q}}_i=\{ \hat{q}_{i, 1}, \hat{q}_{i, 2}, ... \}$, each of which related to a true plausible answer of $\mathcal{A}_i$, we draw inspiration from a human's reasoning chain inferring multiple interpretations of a question. Given an ambiguous question, a human would first identify the ambiguous part of the question and then determine the reason for the ambiguity, followed by inferring multiple interpretations of the question. For example, given the question "Who played the Weasley brothers in Harry Potter?", the ambiguous part is the object of the question, "Weasley brothers," and the corresponding reason is that "It can refer to multiple characters such as Ron, Percy, and so on." Consequently, a human would generate "Who played Ron Weasley in Harry Potter?", "Who played Percy Weasley in Harry Potter?", etc. 

To mimic this reasoning chain, we leverage the LLM's reasoning ability to 1) identify the ambiguous part of the question and the reason for the ambiguity and 2) infer the \textit{pseudo-interpretations} $\hat{\mathcal{Q}}_i$ from the results. But, handling both tasks simultaneously would place a substantial load on a single LLM (See Appendix~\ref{sec:ap-discussion-pseudo-interpretation-inference} for a detailed discussion). As a result, we assign each task to a different LLM, formally represented as:

\vspace{-2ex}
\begin{equation}
\hat{\mathcal{Q}}_i \leftarrow \texttt{LLM}(q_i, I_\text{p}, \texttt{LLM}(q_i, I_\text{a})),
\label{eq:infer-pseudo-interpretations}
\end{equation}
\vspace{-3ex}

\noindent where $I_\text{p}$ and $I_\text{a}$ are carefully designed instructions for each step, respectively. We present the conceptual example of $I_{\text{a}}$ and $I_{\text{p}}$ in Fig~\ref{fig:conceptual_prompt} and full instructions in Table~\ref{tab:ap:prompt_type_reason_infer} and \ref{tab:ap:prompt_pseudo_inter} of Appendix \ref{sec:ap:prompt}. For \texttt{LLM}$(\cdot)$, we consider GPT-3.5 \cite{brown2020language} and GPT-4 \cite{achiam2023gpt}.

\noindent \textbf{Retrieving Relevant and Diverse Passages.} \@  As a first stage retrieval, we obtain the candidate passages $\mathcal{C}_i$ generally relevant to the given question $q_i$ from Wikipedia\footnote{We use ColBERT \cite{khattab2020colbert} and Bing search API as retrievers.}.
From the $\mathcal{C}_i$, we select a set of multiple passages $\mathcal{P}_i$ with maximal coverage of all distinct \textit{pseudo-interpretations} $\hat{\mathcal{Q}}_i$. 

Retrieval for unambiguous questions involves scoring a \textbf{single passage} individually based on their relevance to a \textbf{single interpretation}. Whereas, when it comes to the ambiguous questions, we should retrieve a \textbf{set of passages} encompassing \textbf{multiple interpretations}, which makes this problem more challenging. To obtain such set of passages, we explicitly employ our inferred \textit{pseudo-interpretations} $\mathcal{\hat{Q}}_i$ to retrieve the set of passages $\tilde{\mathcal{P}}_i$ that maximally cover these interpretations, formally represented as follows:

\vspace{-2ex}
\begin{equation}
\tilde{\mathcal{P}}_i \leftarrow \bigcup_{j=1}^{|\mathcal{\hat{Q}}_i|} \mathcal{R}(\mathcal{C}_i, \hat{q}_{i, j}; K),
\label{eq:retrieval-by-pseudo-interpret}
\end{equation}
\vspace{-2ex}

\noindent where $\mathcal{R}$ is a retriever yielding top-$K$ passages from the $\mathcal{C}_i$ by relevance scores to each \textit{pseudo interpretation} $\hat{q}_{i, j}$. 

\noindent \textbf{Pruning Noisy Passages.} \@ Although this process explicitly enables $\mathcal{\tilde{P}}_i$ to encompass all \textit{pseudo interpretations}, there could be some noisy and irrelevant passages due to the absence of perfect retriever and the noise of the inferred \textit{pseudo-interpretations} $\mathcal{\hat{Q}}_i$.  To this end, we find and prune the passages that are highly likely to be irrelevant and noisy. Our intuition is that 1) noisy passages caused by the imperfect retriever tend to be irrelevant to all \textit{pseudo-interpretations} and 2) noisy passages caused by noisy \textit{pseudo-interpretations} tend to be irrelevant to most of the \textit{pseudo-interpretations}. Based upon this intuition, we measure an averaged relevance of the passage to determine if it is noisy or not. The averaged relevance of a passage $\mathcal{S}(p)$ is calculated as follows:

\vspace{-5ex}
\begin{equation}
\mathcal{S}(p) \leftarrow \frac{1}{|\mathcal{\hat{Q}}_i|} 
\sum_{j=1}^{|\mathcal{\hat{Q}}_i|} \frac{\texttt{Enc}(\hat{q}_j) \cdot \texttt{Enc}(p)}{||\texttt{Enc}(\hat{q}_j)|| \cdot ||\texttt{Enc}(p)||},
\label{eq:pruning-noisy-passage}
\end{equation}
\vspace{-2ex}

\noindent where \texttt{Enc}$(\cdot)$ encodes sentences to a dense vector and $p \in \mathcal{\tilde{P}}_i$. We then select the top-$K$ passages from the $\mathcal{\tilde{P}}_i$ based on these averaged scores as the final passage set $\mathcal{P}_i$. 

Our approach is generic, allowing for the use of various sentence embedding models for calculating relevance scores. In line with the sota baseline \cite{kim2023tree}, we employ the frozen SentenceBERT \cite{reimers2019sentence} in $\mathcal{R}(\cdot)$ and \texttt{Enc}$(\cdot)$ in our implementation.

\subsection{Adaptive Generation (AG)}
\label{sec:retrieval-verification}

Despite the effectiveness of the proposed RD method, there may be the low quality of $\mathcal{P}_i$. To further enhance the robustness of \proposed, in this section, we propose an adaptive generation method. The key idea of AG is to carefully verify the overall quality of the passages retrieved from RD before indiscriminately incorporating them. 

From the findings in Section~\ref{sec:preliminary}, if $\mathcal{P}_i$ does not encompass all plausible interpretations, $\mathcal{Q}_i$ and $\mathcal{A}_i$, the response generated by the RAG framework is highly likely to be inaccurate. To this end, we introduce an adaptive generation (AG) method that dynamically adjust the response generation strategy among the RAG framework and closed-book LLM, which is achieved by verifying the quality of $\mathcal{P}_i$ before attempting a solution.

\noindent \textbf{Retrieval Verification (RV)} \@ To verify the quality of $\mathcal{P}_i$, we exploit the LLM’s strong natural language understanding ability. The existing works \cite{li2023llatrieval, asai2023self, yan2024corrective} verify whether $\mathcal{P}_i$ can sufficiently support answering $q_i$ by prompting or training the LLM to give a proper label $V_i$ (e.g., Yes / No): 


\vspace{-2ex}
\begin{equation}
V_i \leftarrow \texttt{LLM}(q_i, \mathcal{P}_i, I_\text{v}),
\end{equation}
\label{eq:aq-verify}
\vspace{-3ex}

\noindent where $I_\text{v}$ is the corresponding instruction. However, the retrieval quality in terms of ambiguous questions should be graded according to how many interpretations are encompassed by the retrieved passages, which can not be achieved by the existing approaches tailored to unambiguous questions. To this end, we newly define a criterion of quality levels tailored to ambiguous questions: $\{\texttt{Useful}, \texttt{PartialUseful}, \texttt{Useless}\}$. \texttt{Useful} indicates the $\mathcal{P}_i$ encompasses all $\mathcal{Q}_i$ and $\mathcal{A}_i$, \texttt{Useless} indicates the $\mathcal{P}_i$ does not contain any of them, otherwise \texttt{PartialUseful}. To determine these grades, we estimate how many interpretations are encompassed by the $\mathcal{P}_i$ by explicitly utilizing the \textit{pseudo-interpretations} $\mathcal{\hat{Q}}_i$:


\vspace{-4ex}
\begin{align}
& V_{i,1} \leftarrow \texttt{LLM}(\hat{q}_{i,1}, \mathcal{P}_i, I_{\text{v}}) \nonumber \\
& \quad\quad\quad\quad\quad\vdots \nonumber \\ 
& V_{i,|\mathcal{\hat{Q}}_i|} \leftarrow \texttt{LLM}(\hat{q}_{i,|\mathcal{\hat{Q}}_i|}, \mathcal{P}_i, I_{\text{v}}),
\label{eq:dq_verify}
\vspace{-3ex}
\end{align}
\vspace{-4ex}

\noindent where each $V_{i,j}$ consists of a binary label (i.e., Yes or No). For instance, if all $V_{i,*}$ are determined "Yes" the grade is \texttt{Useful}. We present the full prompt of $I_{\text{v}}$ in Table~\ref{tab:ap:retrievel_verify} in Appendix~\ref{sec:ap:prompt}. For  \texttt{LLM}$(\cdot)$, we consider GPT-3.5 and GPT-4. 



\noindent \textbf{Adaptive Generation} \@ Once we get the verification results from Eq~\ref{eq:dq_verify}, if the $\mathcal{P}_i$ is classified to \texttt{Useful} or \texttt{PartialUseful}, we decide to utilize the retrieved passages $\mathcal{P}_i$ to generate a response by Eq~\ref{eq:disambiguation} and \ref{eq:question-answering}. If $\mathcal{P}_i$ is classified to \texttt{Useless}, we decide to only utilize the LLM's internal knowledge to generate a response: $\texttt{LLM}(q_i, I_\text{l})$. The full prompt of $I_{\text{l}}$ is presented in Table~\ref{tab:ap:prompt_closedbookllm_response_generation} in Appendix~\ref{sec:ap:prompt}. This process enables the utilization of the most suitable approach tailored to each retrieval quality, which is beneficial to both accuracy and efficiency. 

\subsection{Discussion}
 
\begin{figure}
    \centering
    \includegraphics[width=0.7\columnwidth]{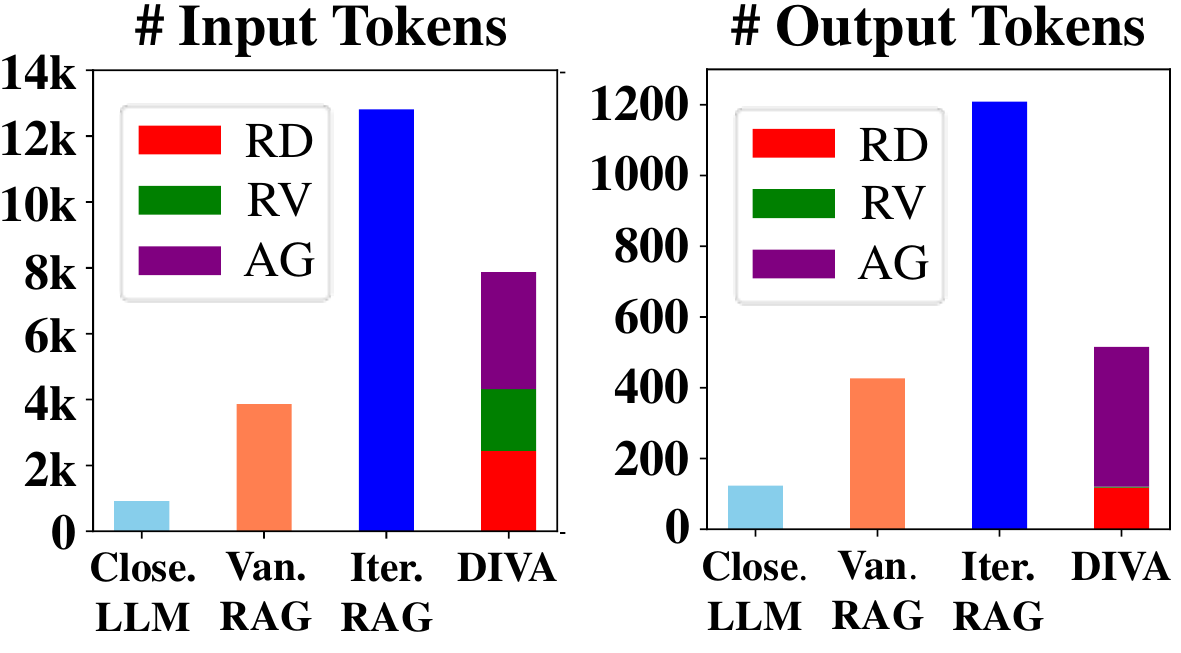}
    \vspace{-2ex}
    \caption{Comparison of the number of tokens per query using GPT-4 backbone. RD, RV, and AG indicate the proposed retrieval diversify, retrieval verify, and adaptive generate module, respectively.
    }
    
    \label{fig:num_tokens} 
    \vspace{-5ex}
\end{figure}

\noindent \textbf{Efficiency.} \@ We examine the factors contributing to \proposed's strong efficiency in Figure~\ref{fig:num_tokens}, which illustrates the average number of input and output tokens per query when using the GPT-4 backbone. For a detailed explanation of the token consumption calculation process, please refer to Appendix~\ref{sec:ap-token-consumption}. \textbf{First, \proposed's strong efficiency is largely due to the RD method.} Unlike Iterative RAG, which involves an average of 5.5 exploration steps per query and requires more than 12,000 tokens for input and 1,200 tokens for output, the RD method significantly reduces the number of tokens needed. \textbf{Second, although the RV method introduces some additional costs, these are acceptable compared to the complexity of Iterative RAG.} Moreover, RV enables the adaptive generation (AG) strategy, where the faster closed-book LLM is selectively used instead of RAG, further enhancing efficiency. {As a result, \proposed, combining RD, RV, and AG, requires substantially less inference time and API costs.}

\noindent \textbf{Technical Contribution.} \@ 
Our paper is the first attempt to investigate how current RAG methods struggle when handling ambiguous queries, and introduces several novel methods specifically tailored to ambiguous queries: retrieval diversification (RD), and retrieval verification (RV). The primary innovation of RD lies in its unique pseudo-interpretation inference, which mimics human reasoning process, and retrieval method. Furthermore, our RV module offers a novel approach that formulates retrieval verification in the context of ambiguous queries. Therefore, our work provides novel insights and strategies targeted at the ambiguous questions, offering a robust and efficient solution to the issues in previous RAG approaches.

\section{Experimental Setups}
\label{sec:experiments-setup}

\subsection{Datasets}
Our proposed method and all baseline models are assessed using the ASQA \cite{stelmakh2022asqa} and SituatedQA \cite{zhang-choi-2021-situatedqa} datasets. ASQA is a long-form QA dataset featuring ambiguous questions. SituatedQA is a short-form QA dataset featuring questions that specifically highlight ambiguities related to temporal and geographical contexts. We give these questions to the QA systems and assess how comprehensively the responses cover the provided possible interpretations of questions. Further details about the datasets are provided in the Appendix~\ref{sec:ap:dataset}.

\subsection{Evaluation Metrics}
\noindent \textbf{Metrics for QA.} \@ Following \citet{min2020ambigqa}, we mainly adopt F1-based metrics. For the short-form QA dataset (SituatedQA) we utilize F1 score. Given ASQA is the long-form QA dataset, following \citet{stelmakh2022asqa}, we use Disambig-F1 (D-F1) score instead of F1. We further leverage ROUGE-L (R-L) to measure correctness of the long-form responses. Finally, Disambiguation-ROUGE (DR), combines R-L and D-F1 scores for overall performance.

\noindent \textbf{Metrics for Passage Retrieval.} \@ Following \citet{min2021joint}, we use MRecall@$k$ to evaluate the quality of retrieved passages. 

For more details of the evaluation metrics, please refer to Appendix \ref{sec:ap:eval-metrics}.

\subsection{Baselines}

We compare our \proposed~against relevant models, including fully-supervised LMs, few-shot closed book LLMs, LLMs w/ RAG, and the adaptive generation. Specifically, fully-supervised LMs include the \textbf{1) T5 closed-book} \cite{raffel2020exploring}, \textbf{2) T5 w/ JPR} \cite{min2021joint}, and \textbf{3)} \textbf{PaLM} \cite{chowdhery2023palm} \textbf{w/ Soft Prompt Tuning}.
Few-shot closed book LLMs include \textbf{4) Vanilla Llama3, GPT-3.5-turbo, and GPT-4}  and \textbf{5) Query refinement} \cite{amplayo2022query}. Few-shot LLMs w/ RAG include \textbf{6) Vanilla RAG} where we use RAC prompt in \citet{kim2023tree}, for \textbf{7) Iterative RAG} we use the sota method ToC \cite{kim2023tree}, for adaptive generation \textbf{8) Self-RAG} \cite{asai2023self}, and for RAG with retrieval verification \textbf{9) CRAG} \cite{yan2024corrective}. For more details of the baselines, please refer to Appendix \ref{sec:ap:baseline}.

\subsection{Implementation Details}
In \proposed, the LLM is employed across three modules: retrieval diversification (Eqn~\ref{eq:infer-pseudo-interpretations}), retrieval verification (Eqn~\ref{eq:dq_verify}), and adaptive response generation (Eqn~\ref{eq:disambiguation}, \ref{eq:question-answering}, and closed-book LLM). For adaptive response generation, we use the same LLM backbones as the other baselines. For the retrieval diversification and verification modules, we assess the performance of GPT-3.5 (gpt-35-turbo) and GPT-4 (gpt-4-0613) across them, ultimately opting to use GPT-4 for both modules in the ASQA dataset and GPT-3.5 for both modules in the SituatedQA dataset in all experiments. However, as demonstrated in Section~\ref{sec:sensi}, other LLMs also perform effectively in these modules. For other implementation details, please refer to Appendix~\ref{sec:ap:implementation-detail}.


%

\section{Experimental Results and Analyses}

\subsection{Main Results}
\label{sec:main-results}

In this section, we assess the effectiveness of \proposed~on ambiguous and unambiguous questions. 

Table~\ref{tab:main_table_asqa} presents the long-form ambiguous QA performance of baselines and \proposed~on the development set of ASQA. 

\textbf{First, \proposed~outperforms the sota baseline, Iterative RAG, in terms of both accuracy and efficiency of response generation.} 
Our method enhances Vanilla RAG framework by incorporating retrieval diversification and adaptive generation strategies that address low-quality retrieval and improve performance. It is also more efficient, requiring significantly less computational overhead and achieving 1.5x - 3x greater efficiency in inference time across various LLM backbones compared to Iterative RAG. Overall, our method produces more accurate and diverse interpretations without the cumbersome iterative exploration process.

\textbf{Second, \proposed~outperforms the recent retrieval verifying method, CRAG, across all metrics, including inference time.} CRAG underperforms even compared to Vanilla RAG, despite its verification and correction mechanisms. This suggests that CRAG’s verification and correction methods are not well-suited to handling ambiguous queries, resulting in degraded passage retrieval performance. These findings emphasize the need for a RAG method specifically designed for ambiguous queries, demonstrating the practicality and effectiveness of \proposed~in such scenarios.

\textbf{Third, \proposed~demonstrates good adaptability in switching out the underlying LLM backbones.} 
\proposed~consistently enhances Vanilla RAG with its RD and AG modules across different LLM backbones regardless of their model sizes (Llama3-8B to GPT-4), demonstrating its adaptability and wide applicability. This suggests that \proposed~can easily integrate with more advanced LLMs in the future.

\begin{figure}
    \centering
    \includegraphics[width=0.9\columnwidth]{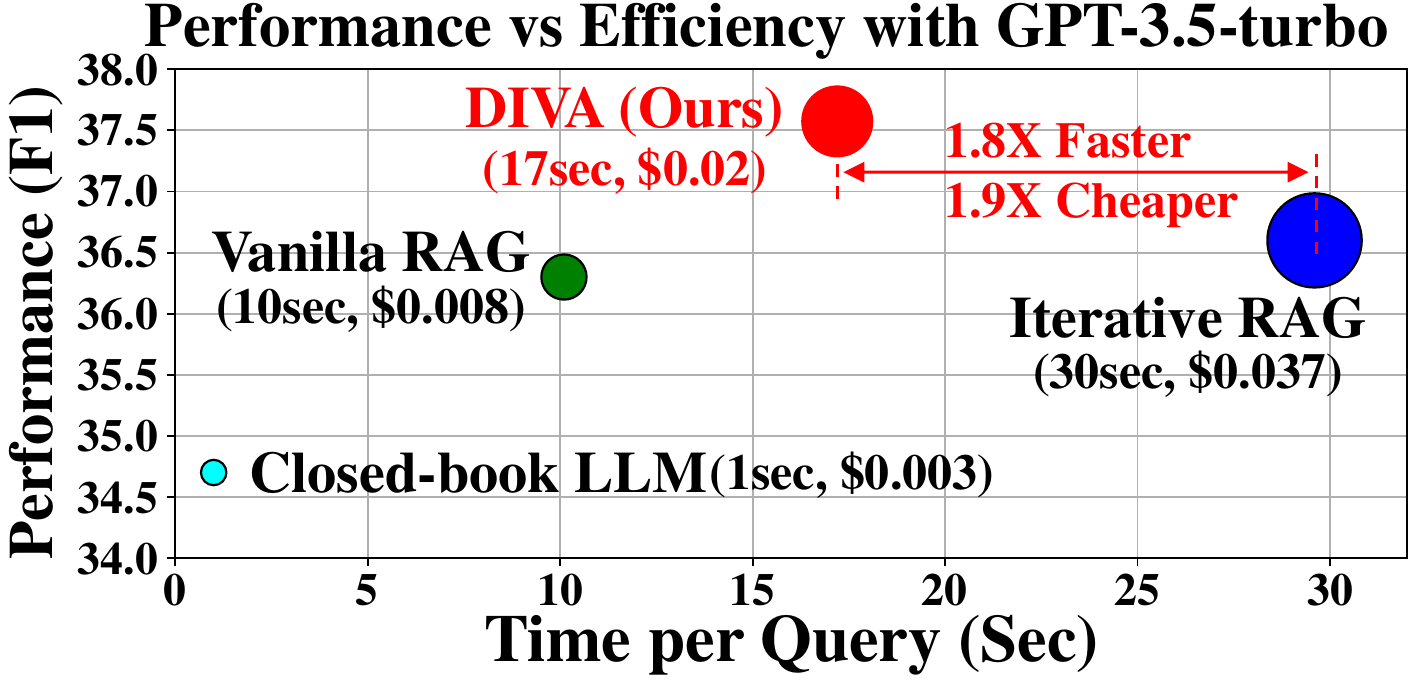}
    \vspace{-2ex}
    \caption{Experiments on SituatedQA dataset. 
    }
    
    \label{fig:main-situ} 
    \vspace{-3ex}
\end{figure}

Fig~\ref{fig:main-situ} shows the performance and efficiency of baselines and \proposed~on the SituatedQA test set for short-form ambiguous QA tasks. All experimental results align with those seen in Table~\ref{tab:main_table_asqa}, demonstrating strong generalizability of \proposed~across different types of ambiguous questions.

\textbf{Finally, \proposed~exhibits strong performance on unambiguous questions as well}, i.e., NQ dataset \cite{kwiatkowski-etal-2019-natural}, highlighting its broad applicability. For detailed results and explanations, please refer to Appendix~\ref{sec:unambiguous-experiments}.

\begin{table}[ht]
    \centering

    \resizebox{1.\columnwidth}{!}{\begin{tabular}{lcccc}
        \toprule
        & \multicolumn{1}{c}{\textbf{R-L}} & \multicolumn{1}{c}{\textbf{D-F1}} & \multicolumn{1}{c}{\textbf{DR}} & \textbf{Time}  \\
        \midrule
        \rowcolor{gray!20}\multicolumn{5}{c}{\textbf{Fully-Supervised}} \\
        T5-Large Closed-Book\textsuperscript{$\star$} & 33.5 & 7.4 & 15.7 & - \\
        T5-Large w/ JPR\textsuperscript{$\star$} & {43.0} & 26.4 & 33.7 & -  \\
        PaLM w/ Soft Prompt Tuning\textsuperscript{$\star\star$} & 37.4 & 27.8 & 32.1 & - \\

        \rowcolor{gray!20}\multicolumn{5}{c}{\textbf{Few-shot Prompting: Closed-Book LLM}}\\
        Llama3-8B-Instruct & 31.1	& 25.6	& 28.2 & - \\
        Llama3-70B-Instruct & 35.7	& 36.4	& 36.0 & 30.5\\
        \midrule
        GPT-3.5-turbo & 38.8	& 34.0 & 36.3 & 2.0 \\
        $+$ Query Refinement & 37.5 & 34.8 & 36.1 & 5.2 \\
        \midrule
        GPT-4 & 39.0	& 38.5 & 38.7 & 5.9 \\
        $+$ Query Refinement & 39.6 & 39.3 & 39.4 & 10.0 \\

        \rowcolor{gray!20}\multicolumn{5}{c}{\textbf{Few-shot Prompting: LLM w/ RAG}}\\
        Self-RAG-13B  & 35.4 & 26.0 & 30.4 & 4.1 \\
        CRAG (GPT-4) & 40.1 & 39.6 & 39.9 & 34.4 \\
        \midrule
        \textbf{Llama3-8B-Instruct} & \\
        --- \textit{Vanilla RAG} & 38.2	& 35.4 &	36.8 & - \\
        --- \textit{Iterative RAG (ToC)}  & 37 & 36.3 & 36.6 & - \\
        --- \textit{\proposed~(Ours)} & \textbf{38.9}	& \textbf{35.7} &	\textbf{37.3}  & - \\
        
        \midrule
        \textbf{Llama3-70B-Instruct} & \\
        --- \textit{Vanilla RAG} & 40.2	& 40.0 &	40.1 & 42.3\\
        --- \textit{Iterative RAG (ToC)}  & 39.5 & 40.4 & 39.9 & 140.5 \\
        
        --- \textit{\proposed~(Ours)} & \textbf{40.4}	& \textbf{41.4} &	\textbf{40.9}  & \textbf{50.6} \\
        \midrule
        \textbf{GPT-3.5-turbo} & \\
        --- \textit{Vanilla RAG} & 41.2	& 37.5	& 39.3 & 11.2\\
        --- \textit{Iterative RAG (ToC)}  & 40.1	& 38.5 &	39.3 & 31.5 \\
        
        --- \textit{\proposed~(Ours)} & \textbf{42.1}	& \textbf{38.9} &	\textbf{40.5} & \textbf{19.8} \\
        \midrule
        \textbf{GPT-4} & \\
        --- \textit{Vanilla RAG} & 41.5	& 39.6 & 40.6 & 18.9\\
        --- \textit{Iterative RAG (ToC)}  & 38.5 & 41.8 & 40.1 & 53.1 \\
        
        --- \textit{\proposed~(Ours)} & \textbf{42.4}	& \textbf{42.0}	& \textbf{42.2} & \textbf{27.1} \\
        \bottomrule 
        \multicolumn{4}{l}{$\star$ results from \citet{stelmakh2022asqa}} \\
        \multicolumn{4}{l}{$\star\star$ results from \citet{amplayo2022query}} \\

        
    \end{tabular}}
    \vspace{-2ex}
        \caption{Experiments on ASQA dataset.  Baselines are either fully-supervised or 5-shot prompted. The metric Time indicates inference time (sec) per query. 
        We emphasize our results in bold, for easy comparisons.}
    \label{tab:main_table_asqa}
\end{table}

\subsection{Ablation Studies}

\begin{table}[ht]
    
    \centering
    \resizebox{1.\columnwidth}{!}{\begin{tabular}{c|ccc|ccc|ccc}
        \toprule
        & \multicolumn{3}{c|}{\textbf{Component}} & \multicolumn{3}{c}{\textbf{GPT-3.5-turbo}} & \multicolumn{3}{|c}{\textbf{GPT-4}}\\
        
        \textbf{Row} & \textbf{RAG} & \textbf{RD} & \textbf{AG}  & \multicolumn{1}{c}{\textbf{R-L}} & \multicolumn{1}{c}{\textbf{D-F1}} & \multicolumn{1}{c}{\textbf{DR}} & \multicolumn{1}{|c}{\textbf{R-L}} & \multicolumn{1}{c}{\textbf{D-F1}} & \multicolumn{1}{c}{\textbf{DR}} \\
        \midrule
        1 & \ding{55} & \ding{55} & \ding{55} & 38.8 & 34.0	& 36.3 & 39.0	& 38.5 &	38.7 \\
        2 & \ding{51} & \ding{55} & \ding{55} & \underline{41.2} &	37.5 &	39.3 & 41.5	& 39.6	& 40.6 \\
        3 & \ding{51} & \ding{51} & \ding{55}  & \textbf{42.1}	& \underline{38.5} &	\underline{40.2} & \underline{42.3} &	41.0 &	\underline{41.7}  \\
        4 & \ding{51} & \ding{51} & \ding{51}  & \textbf{42.1}	& \textbf{38.9} & \textbf{40.5} & \textbf{42.4} & 	\textbf{42.0} & \textbf{42.2} \\
        \bottomrule        
    \end{tabular}}
    \vspace{-2ex}
    \caption{Ablation studies on ASQA dataset. }
    \label{tab:ablation_asqa}
\end{table}

To evaluate the importance of each component of \proposed, namely retrieval diversification (RD) and adaptive generation (AG), we incrementally add them to Vanilla RAG (row 2 in Table~\ref{tab:ablation_asqa}). Table~\ref{tab:ablation_asqa} reveals the following insights: 
\textbf{1)} RAG (row 2) with the closed-book LLM (row 1) significantly enhances the ability to handle ambiguity in questions. 
\textbf{2)} Implementing the RD module (row 3) enhances all performance metrics, demonstrating that RD effectively diversifies and improves the quality of retrieved passages, thereby enhancing the RAG framework. \textbf{3)} Incorporating the AG module (row 4) also boosts all metrics, showing that the retrieval verification method accurately identifies \texttt{Useless} passages. Additionally, this supports our finding in Sec~\ref{sec:preliminary} that when retrieved passages are of extremely low quality, the internal knowledge of LLMs proves more advantageous than RAG.

\subsection{Retrieval Analysis}

\begin{table}[ht]
    \centering
    \resizebox{1.0\columnwidth}{!}{\begin{tabular}{c|c|c|cc}
        \toprule
         & & \multicolumn{1}{c}{\textbf{MRecall@$k$}} & \multicolumn{2}{|c}{\textbf{D-F1}}\\
        
        \textbf{Row} & \textbf{Method} & \multicolumn{1}{c}{\textbf{$k=5$}} & \multicolumn{1}{|c}{\textbf{GPT-3.5-turbo}} & \multicolumn{1}{c}{\textbf{GPT-4}} \\
        \midrule
        1 & Vanilla RAG & 35.2 &	37.5 &	39.6 \\
        \midrule
        2 & $+$ \citet{ma2023query} & 36.1 &	37.0 &	40.4 \\
        3 & $+$ RD (Ours) & \textbf{37.0} &	\textbf{38.5} &	\textbf{41.0} \\
        4 & $+$ Oracle & 41.5 & - & - \\
        \bottomrule        
    \end{tabular}}
    \caption{Retrieval accuracy and corresponding QA performance on ASQA dataset. }
    \label{tab:retrieval}
\end{table}

We evaluate the effectiveness of our proposed RD method in Table~\ref{tab:retrieval} using MRecall@$k$ \cite{min2021joint}. Vanilla RAG (row 1) involves basic retrieval of passages using a given question $q_i$. "$+$ RD" (row 3) applies the RD method to row 1, using \textit{pseudo-interpretations} generated by our proposed instructions (i.e., $I_\text{p}$ and $I_\text{a}$). Row 2 uses the RD method with \textit{pseudo-interpretations} generated by the LLM query rewriter as described in \citet{ma2023query} using simple instructions. "$+$ Oracle" (row 4) applies RD to Vanilla RAG using ground-truth interpretations from the ASQA dataset.

We observe that \textbf{1)} adding RD leads to significant improvements of MRecall and D-F1 score compared to Vanilla RAG, demonstrating RD effectively addresses low-quality retrieval issue and then improve the QA performance. \textbf{2)} "$+$ RD" outperforms "$+$ \citet{ma2023query}" showing the superiority of our carefully designed instruction in inferring \textit{pseudo-interpretations}. \textbf{3)} "$+$ Oracle" (row 4) significantly outperforms RD, indicating that when more advanced LLMs are available in the future there is potential for RD to improve in accurately inferring \textit{pseudo-interpretations}.

\vspace{-1ex}
\subsection{Sensitivity Analysis}
\label{sec:sensi}

\begin{figure}
    \centering
    \includegraphics[width=\columnwidth]{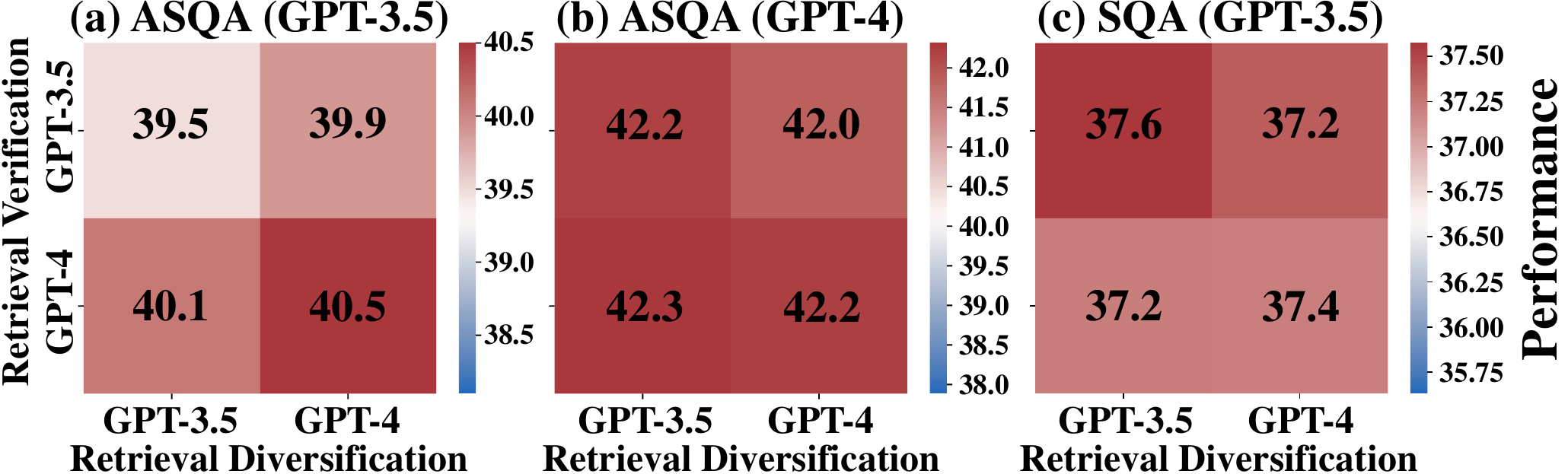}
    \vspace{-4ex}
    \caption{Sensitivity analysis of LLM backbone model in retrieval diversification and verification modules. \textcolor{red}{Red}-white-\textcolor{blue}{blue} means outperformance, on-par, and underperformance compared with Iterative RAG (ToC) in terms of DR for ASQA and F1 for SituatedQA.
    }
    
    \label{fig:sensi} 
    \vspace{-3ex}
\end{figure}

For the retrieval diversification (RD) and retrieval verification (RV) modules, we explore how their performance is affected by the choice of LLM. We evaluate the impact of using GPT-3.5 and GPT-4 across both modules, comparing the overall QA performance against the sota baseline, ToC \cite{kim2023tree}, on the ASQA and SituatedQA datasets. Fig~\ref{fig:sensi}(a) and (b) represent using GPT-3.5 and GPT-4 as the response generation models on the ASQA dataset, respectively. Fig~\ref{fig:sensi}(c) represents using GPT-3.5 as the response generation model on the SituatedQA dataset.

In Fig~\ref{fig:sensi}, we observe: \textbf{1)} \proposed~consistently outperforms ToC, regardless of the LLM model used in each module. \textbf{2)} While the RD module shows very stable results, the RV module appears relatively sensitive to the choice of LLM. This highlights that verifying the quality of retrieved passages for ambiguous questions requires more powerful natural language understanding ability, underscoring the need for future work to alleviate the dependency on the choice of LLM. Based on these results, we argue that \proposed~is a general framework that is robust across different LLMs.

\vspace{-1ex}
\subsection{Case Study}

\begin{figure}
    \centering
    \includegraphics[width=\columnwidth]{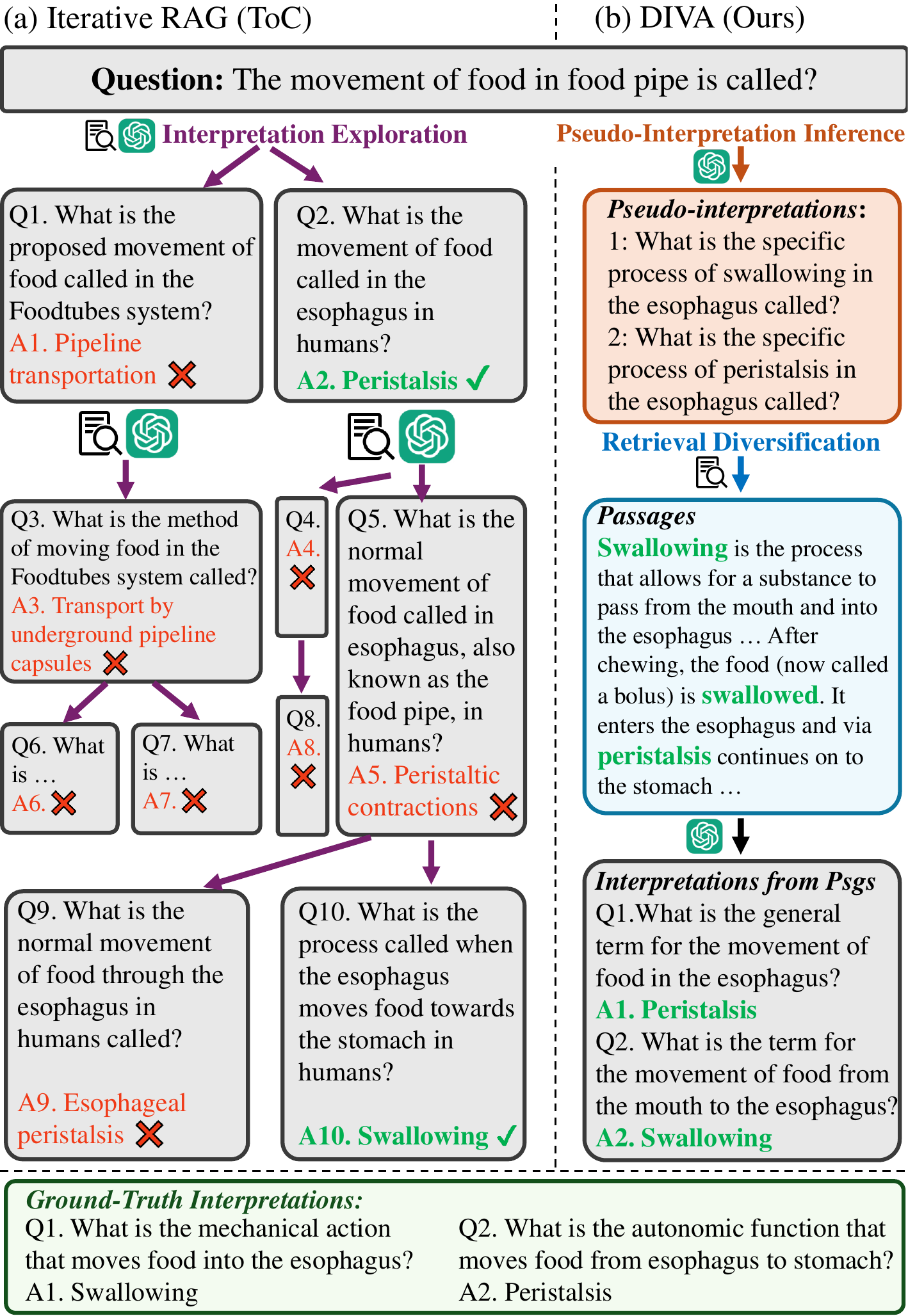}
    \caption{Case study with GPT-4.} 
    \label{fig:case-study1} 
    \vspace{-2ex}
\end{figure}

We conduct a case study to qualitatively compare the reasoning chains of Iterative RAG, ToC \cite{kim2023tree}, and \proposed. Figure~\ref{fig:case-study1} illustrates the reasoning chains of Iterative RAG, ToC \cite{kim2023tree}, and \proposed~using the ASQA question, "The movement of food in the food pipe is called?". In panel (a), the answer "Peristalsis" is easily covered during the first exploration, whereas "Swallowing" requires six steps of passage retrieval and LLM reasoning for exploration. In contrast, panel (b) shows that our \textit{pseudo-interpretations} include both interpretations, with the RD retrieving passages that encompass all necessary information. Consequently, the LLM efficiently extracts all plausible interpretations from the retrieved passages without the need for the cumbersome iterative exploration process.

Moreover, we analyze failure cases to provide valuable insights into the limitations and potential improvements of \proposed~by identifying instances where it underperforms. Detailed results and explanations are presented in Appendix~\ref{sec:ap:failure-cases}.

\vspace{-1ex}
\section{Related Work}
\vspace{-1ex}
\label{sec:relatedwork}

\noindent \textbf{RAG for Ambiguous Question.} \@ 
To tackle the ambiguity inherent in certain questions, earlier studies \cite{min2021joint, gao2020answering, shao2021answering, iterativeprompting} necessitated the fine-tuning of models using extensive training datasets. 
Recently, some studies have leveraged LLM to generate comprehensive responses through an in-context learning. For example, RAC \cite{kim2023tree} instructs LLM to extract plausible interpretations and answers from provided passages. However, they overlook the problem of low-quality retrieval, which results in significant performance drops. To tackle this issue, ToC \cite{kim2023tree} explores missing interpretations by an iterative passage retrieval and LLM reasoning. However, the iterative process incurs significant computational overhead. 

\noindent \textbf{Query Reformulation.} \@ 
Our pseudo-interpretation inference may appear to share similarities with existing query reformulation approaches. The first line of research is query decomposition \cite{min2019multi, khot2022decomposed}, which breaks down complex queries containing multi-hop relations or an overabundance of information. In contrast, our work addresses the opposite challenge, ambiguity, caused by a lack of information within the query, rather than the overabundance of information. Therefore, query decomposition and pseudo-interpretation inference are fundamentally different in purpose.
Another line is query rewriting. \citet{ma2023query} utilize an LLM and naively designed prompt to rewrite the given query to improve the quality of retrieval. Compared to \citet{ma2023query}, the primary innovation of \proposed~lies in its unique and well-designed prompting method, which imitates human reasoning chains to infer the pseudo-interpretations from ambiguous queries. 



\noindent \textbf{Retrieval Quality Verification.} \@ 
Many studies have noted that low-quality retrieval introduces significant irrelevant information to the RAG framework and have proposed various solutions. Self-RAG \cite{asai2023self} fine-tunes LLM to generate a reflection token that assesses the relevance of a passage to the question at hand. Llatrieval \cite{li2023llatrieval} employs LLM to check if retrieved passages sufficiently support the answer, updating them if they are of low quality. Meanwhile, CRAG \cite{yan2024corrective} trains a lightweight verifier to evaluate the quality of retrieved passages, making corrections if they fall below a set threshold. Please note that our work focuses on retrieval verification, distinguishing it from other methods such as CoVe \cite{dhuliawala2023chain} and Verify-and-Edit \cite{zhao2023verify}, which do not target this aspect. For a more detailed discussion, please refer to Appendix~\ref{ap:comp-verfication}.



\noindent \textbf{Adaptive Generation.} \@ 
Numerous studies have examined adaptive strategies that dynamically determine the need for retrieval, utilizing only the internal knowledge of LLMs when unnecessary \cite{mallen-etal-2023-trust, feng2023trends}. \citet{mallen-etal-2023-trust} used an empirical method to retrieval, activating relying on the frequency of entity. 
AdaptiveRAG \cite{jeong2024adaptive} dynamically chooses the optimal response generation strategy tailored to the complexity of the query. TA-ARE \cite{zhang2024retrievalqa} uses in-context learning to assess whether a query necessitates retrieval. 

Compared with recent studies that either overlook or inefficiently address the issue of low-quality retrieval in ambiguous questions, we introduce the retrieval diversification method efficiently retrieves higher quality passages without relying on cumbersome iterative processes. Additionally, we propose retrieval verification and adaptive generation strategies specifically designed for ambiguous questions, while the existing works overlook these important challenge of ambiguous questions. To the best of our knowledge, this paper is the first effort to thoroughly analyze and address the problem of low-quality retrieval in the context of ambiguous questions and its potential solutions.

\vspace{-1ex}
\section{Conclusion}
\vspace{-1ex}
\label{sec:conclusion}
In this study, we examined the shortcomings of the current RAG-based method in dealing with ambiguous questions, specifically its low-quality retrieval and inefficiency. Our proposed framework, \proposed~, effectively diversifies the retrieved passages to capture various interpretations, verifies their quality, and adapts the most appropriate approach based on that quality. This strategy improves QA performance while minimizing inefficiency.

\section*{Limitations}

While \proposed~demonstrates clear advantages in effectiveness and efficiency through retrieval diversification and adaptive generation, its design is specifically tailored for ambiguous questions. In real-world QA systems, where queries can be a mix of ambiguous and unambiguous, the applicability of \proposed~may be limited. However, recent work has introduced methods to classify whether a query is ambiguous \cite{cole2023selectively}, which leads to utilizing the suitable approach according to its ambiguity. Although \citet{cole2023selectively} proposed simple approaches, there is still significant potential to enhance these methods using advanced techniques like in-context learning and RAG. Future research could focus on developing systematic approaches for classifying the ambiguity of queries. 
Furthermore, the performance of our proposed retrieval verification module is somewhat sensitive to the choice of LLM. Specifically, it tends to work better with GPT-4 than with GPT-3.5, though this may negatively impact the efficiency of \proposed. Therefore, future work should focus on developing a more efficient and robust retrieval quality verifier LLM, tailored to handling ambiguous questions, to enhance both effectiveness and efficiency.


\section*{Ethics Statement}
Given that \proposed~is built on the RAG framework of QA systems, it is important to consider the following points: (1) the retrieved passages may contain offensive or harmful content, which could result in similarly harmful responses, and (2) user queries themselves may be offensive or harmful. Therefore, developing methods to detect harmful user queries and selectively retrieve passages that are free from harmful content could be a crucial focus for future research.

\section*{Acknowledgements}
This work was supported by Institute of Information \& Communications Technology Planning \& Evaluation(IITP) grant funded by the Korea government(MSIT) (RS-2023-00216011, Development of artificial complex intelligence for conceptually understanding and inferring like human) and National Research Foundation of Korea(NRF) funded by Ministry of Science and ICT (NRF-2022M3J6A1063021).
\bibliography{custom}

\newpage

\appendix

\begin{algorithm}[t]
\caption{diversify-verify-adapt (\proposed)}
\begin{algorithmic}[1]
    \Input Question $q_i$, large language model $\texttt{LLM}(\cdot)$, the retriever $\mathcal{R}(\cdot)$, candidate passages $\mathcal{C}_i$
    \Output generated response $r_i$
    \State $\mathcal{\hat{Q}}_i$ $\leftarrow$ Use $\texttt{LLM}(\cdot)$ function to infer the \textit{pseudo-interpretations} by Eq.\ref{eq:infer-pseudo-interpretations}
    
    \State $\mathcal{\tilde{P}}_i$ $\leftarrow$ Use $\mathcal{R}(\cdot)$ function to retrieve relevant and diverse passages by Eq.\ref{eq:retrieval-by-pseudo-interpret}

    \State $\mathcal{S}_i \leftarrow \{\}$
    \For{$j=1$ to $|\mathcal{\tilde{P}}_i|$}
        \State $S(\tilde{p}_{i, j}) \leftarrow$ Obtain noise score by Eq.\ref{eq:pruning-noisy-passage} 
        \State $\mathcal{S}_i \leftarrow \mathcal{S}_i \cup S(\tilde{p}_{i, j})$
    \EndFor
    
    \State $\mathcal{P}_i \leftarrow$ Select top-$K$ passages from $\mathcal{\tilde{P}}_i$ based on the score $\mathcal{S}_i$
    \State $\mathcal{V}_i \leftarrow \{ \}$
    \For{$j=1$ to $|\mathcal{\hat{Q}}_i|$}
        \State $V_{i, j}$ $\leftarrow$ Use $\texttt{LLM}(\cdot)$ function to verify $\mathcal{P}_i$ to $\hat{q}_{i, j}$ by Eq.\ref{eq:dq_verify}
        \State $\mathcal{V}_i \leftarrow \mathcal{V}_i \cup V_{i, j}$ 
    \EndFor

    \If{$\mathcal{V}_i$ is \texttt{Useful} or \texttt{PartialUseful}}
        \State $r_i \leftarrow$ Generate response using $\mathcal{P}_i$ by Eq.\ref{eq:disambiguation} and \ref{eq:question-answering}
    \Else
        \State $r_i \leftarrow \texttt{LLM}(q_i, I_\text{l})$ 
    \EndIf
    \State \textbf{return} $r_i$
\end{algorithmic}
\label{alg:algo1}
\end{algorithm}

\section{Experimental details}

\subsection{Datasets}
\label{sec:ap:dataset}
Our proposed method and all baseline models are assessed using the ASQA \cite{stelmakh2022asqa} and SituatedQA \cite{zhang-choi-2021-situatedqa} datasets. ASQA is a long-form QA dataset derived from a subset of ambiguous questions in the AmbigNQ dataset \cite{min2020ambigqa}. The ASQA dataset contains 6,316 ambiguous questions and their corresponding comprehensive long-form answers that contain all plausible answers, split into 4,353 for training, 948 for development, and 1,015 for testing. SituatedQA is a short-form QA dataset featuring questions that specifically highlight ambiguities related to temporal and geographical contexts. In this dataset, each question is subject to multiple interpretations, with corresponding answers varying by context. We give these questions to the QA systems and assess how comprehensively the responses cover the possible interpretations.

\subsection{Evaluation Metrics}
\label{sec:ap:eval-metrics}

\noindent \textbf{Metrics for QA.} \@ For both datasets, following previous studies on ambiguous QA \cite{min2020ambigqa}, we
mainly adopt F1-based metric. Specifically, for the short-form QA dataset (SituatedQA) we measure F1 based on the precision and recall between the ground-truth answers and the generated responses. Given ASQA is the long-form QA dataset, following \citet{stelmakh2022asqa}, we use Disambig-F1 (D-F1), which assesses the factual accuracy of long-form responses, instead of F1. Using a RoBERTa model \cite{liu2019roberta} trained on SQuAD2.0, we extract short answers from the generated long-form responses and compare them to the ground-truth disambiguation questions (DQs). The F1 score of these extracted answers indicates whether the long-form answers contain correct information. We further leverage ROUGE-L (R-L) to measure correctness of the generated long-form responses to the ground-truth long-form answers. Finally, Disambiguation-ROUGE (DR), combines R-L and D-F1 scores as a geometric mean for overall performance.

\noindent \textbf{Metrics for Passage Retrieval.} \@ Following \citet{min2021joint}, we use MRecall@$k$ to evaluate the quality of retrieved passages by considering retrieval to be successful if all answers or at least $k$ answers in the plausible answer set are recovered by the retrieved passages.

\subsection{Baselines}
\label{sec:ap:baseline}

For all baselines and \proposed, due to the significant costs associated with evaluating RAG models, we perform experiments with a single run.

We describe the details of models as follows:

\noindent \textbf{1) T5 closed-book.} \@ \citet{stelmakh2022asqa} fine-tuned T5-large \cite{raffel2020exploring} to generate long-form response on the whole train set.

\noindent \textbf{2) T5 w/ JPR.} \@ \citet{stelmakh2022asqa} fine-tuned T5-large \cite{raffel2020exploring} with JPR \cite{min2021joint}, fully trained dense retriever for ambiguous QA, to generate long-form response on the whole train set. 

\noindent \textbf{3) PaLM w/ Soft Prompt Tuning.} \@ \citet{amplayo2022query} employed a prompt engineering method to PaLM \cite{chowdhery2023palm} that learn the soft prompts in the closed-book setup. 

\noindent \textbf{4) Closed-book LLM.} \@  Closed-book LLM indicates the traditional few-shot prompting method used in \citet{brown2020language}. We consider the backbone LLM as Llama3-70B-Instruct, GPT-3.5, and GPT-4.

\noindent \textbf{5) Query refinement.} \@ Inspired by \citet{amplayo2022query}, we developed an in-context learning method within a closed-book setup. First, we prompt the LLM to refine ambiguous questions into multiple possible interpretations. These interpretations are then used as in-context examples for the LLM to generate a response that addresses all potential interpretations. We consider the backbone LLM as Llama3-70B-Instruct, GPT-3.5, and GPT-4.

\noindent \textbf{6) Vanilla RAG.} \@ In this method, we begin by retrieving the top 5 relevant passages based on the frozen SentenceBERT similarity between the given query and candidate passages from Wikipedia. We then use the RAC prompt from \citet{kim2023tree} to extract interpretations and generate corresponding answers. We consider the backbone LLM as Llama3-70B-Instruct, GPT-3.5, and GPT-4.

\noindent \textbf{7) Iterative RAG.} \@ For this approach, we employ the state-of-the-art method ToC \cite{kim2023tree} for handling ambiguous QA. Specifically, ToC iteratively constructs a tree of possible interpretations for the ambiguous question using few-shot prompting that leverages external knowledge, and then uses this tree to generate a long-form response. Following the authors' implementation, we set the tree's maximum depth to 3 and the maximum number of nodes to 10. It is important to note that we do not use the tree pruning method in our implementation, as we observe that adding this method notably degrades the QA performance. The retrieval settings are identical to those used in Vanilla RAG. We consider the backbone LLM as Llama3-8B-Instruct, Llama3-70B-Instruct \cite{dubey2024llama}, GPT-3.5, and GPT-4.

\noindent \textbf{8) Self-RAG.} \@ The LLM is trained to adaptively manage retrieval and generation, initiating retrieval when a special token is predicted above a certain threshold, followed by generating the answer. We consider the model trained on Llama2-13B.

\noindent \textbf{9) CRAG.} \@ While the original implementation of CRAG utilized Llama-2-7B, we use GPT-4 as the backbone LLM for a fair comparison, ensuring consistency with the \proposed~setup. In the CRAG implementation, we first retrieve relevant passages for a given query, following the same procedure as Vanilla RAG. Next, we apply CRAG’s retrieval verification and correction procedures to refine these passages. The corrected passages are then fed into GPT-4 using the same instructions as in the Vanilla RAG framework to generate the final response to the query.

\subsection{Implementations Details}
\label{sec:ap:implementation-detail}

Since \proposed~utilizes few-shot prompting, we dynamically select $k$-shot examples through nearest neighbor search and incorporate them into the prompt, following the approach in \citet{kim2023tree} using dsp package \cite{khattab2022demonstrate}. For the retrieved passages $\mathcal{P}_i$, we set the number of passages $|\mathcal{P}_i|$ to 5. We use GPT-4 for both the retrieval diversification and verification steps. For adaptive response generation, we use the same LLM backbones as the other baselines. For the retrieval diversification and verification modules, we assess the performance of GPT-3.5 (gpt-35-turbo) and GPT-4 (gpt-4) across them, ultimately opting to use GPT-4 for both modules in the ASQA dataset and GPT-3.5 for both modules in the SituatedQA dataset for all experiments.  The APIs provided by Microsoft Azure\footnote{https://azure.microsoft.com/} are employed for GPT-3.5-turbo and GPT-4, with the following settings: max tokens set to 300, top-$p$ to 1.0, and temperature to 0.3.

\subsubsection{Retrieval Process}
\label{sec:ap:retrieval-process}
To retrieve relevant passages for the given question, we follow the method utilized by \citet{kim2023tree}. Specifically, we first gather relevant Wikipedia documents for the question using two retrieval systems: ColBERT \cite{khattab2020colbert} and the Bing search engine\footnote{https://www.microsoft.com/bing}. After compiling a set of passages, we rerank and select the top-k passages. For reranking, we utilize SentenceBERT \cite{reimers2019sentence}, pre-trained on MS-Marco, as the backbone. 

\subsubsection{Token Consumption Calculation}
\label{sec:ap-token-consumption}

In this subsection, we explain the overall process of each method and the token consumption calculation on GPT-4. Notably, the token count per query was determined by averaging the input and output tokens across all test queries.

\textbf{Vanilla RAG} is formally described as a sequence of $\texttt{LLM}$ functions as shown in Eq~\ref{eq:disambiguation} and \ref{eq:question-answering}. In Eq~\ref{eq:disambiguation}, given the relevant $\mathcal{P}_i$, the LLM is first prompted with the question $q_i$ and $\mathcal{P}_i$ to extract all plausible interpretations $\mathcal{Q}_i$ and their corresponding answers $\mathcal{A}_i$. During this function call, the average number of input tokens is 1,902, while the average number of output tokens is 177. Note that the input text include task description, few-shot demos, retrieved passages, and given questions, leading to substantial token consumption. Next, in Eq~\ref{eq:question-answering}, based on the $\mathcal{Q}_i$ and $\mathcal{A}_i$, the LLM is prompted to consolidate them with $q_i$ and $\mathcal{P}_i$ to generate a response $r_i$. In this function call, the averaged input tokens are 1,963, and the output tokens are 249. \textbf{Consequently, for Vanilla RAG, the total token consumption is 3,865 input tokens and 426 output tokens.}
    
\textbf{Iterative RAG} is formally represented as iterative $\texttt{LLM}$ function calls in Eq~\ref{eq:disambiguation}, followed by a single $\texttt{LLM}$ call in Eq~\ref{eq:question-answering}. Specifically, after obtaining multiple plausible interpretations $\mathcal{Q}_i$ and their answers $\mathcal{A}_i$ from Eq~\ref{eq:disambiguation}, each interpretation in $\mathcal{Q}_i$ is used for an additional \texttt{LLM} call in Eq~\ref{eq:disambiguation}. This process is repeated, constructing a tree-like structure, until the stopping criterion is met. On average, during this iterative process, the total number of \texttt{LLM} calls converges to 5.5. As a result, the average number of input and output tokens during this process are 10,627 and 936, respectively. Next, same as Vanilla RAG in Eq~\ref{eq:question-answering}, based on all $\mathcal{Q}_i$ and $\mathcal{A}_i$ collected from the iterative process, the LLM is prompted to consolidate them with $q_i$ and $\mathcal{P}_i$ to generate a response $r_i$. In this \texttt{LLM} call, the average number of input tokens is 2,187, and the output tokens is 271. \textbf{Consequently, for Iterative RAG, the total token consumption is 12,814 input tokens and 1,207 output tokens.}

For \textbf{\proposed}, in addition to the calls required by Vanilla RAG, additional LLM calls are introduced through the operations of the RD (Retrieval Diversification) and RV (Retrieval Verification) modules. These additional operations add to the overall token usage. More specifically, in RD, we require two LLM calls to infer a set of multiple pseudo-interpretations, where the number of input and output tokens are 2,440 and 117, respectively. Following this, the retrieval of relevant passages based on the inferred pseudo-interpretations does not require any LLM calls. Verifying the set of retrieved passages requires the same number of LLM calls as the number of pseudo-interpretations. It is important to note that the retrieved passages are concatenated into a single passage before the verification step, allowing for an efficient LLM call process for each pseudo-interpretation. For the verification step, the number of input and output tokens are 1,878 and 3, respectively. It is important to note that if the verifier determines the retrieved passages are not useful, the response is generated using a Closed-book LLM instead of Vanilla RAG. The token consumption for the Closed-book LLM is significantly lower, with 913 input tokens and 123 output tokens. \textbf{Consequently, for \proposed, the total token consumption is 7,873 input tokens and 515 output tokens.}

\begin{table}[ht]
    \centering
    \resizebox{0.9\columnwidth}{!}{\begin{tabular}{c|cc}
        \toprule
         \textbf{Method} & \textbf{\# Input Tokens} & \textbf{\# Output Tokens} \\
        \midrule
         Closed-book LLM & 913 & 123 \\
         Vanilla RAG & 3,865 & 426 \\
        Iterative RAG & 12,814 & 1,207 \\
        \midrule
        \proposed & 7,873 & 515 \\
        
        \bottomrule        
    \end{tabular}}
    \caption{The number of input and output token consumption of each method. }
    \label{tab:token_consumption_table}
\end{table}

\section{Additional Discussion}

\subsection{Regarding the Pseudo-interpretation Inference}
\label{sec:ap-discussion-pseudo-interpretation-inference}

We initially experimented with combining the ambiguity detection and pseudo-interpretation inference into a single step to simplify the process. Specifically, we provided GPT-4 with newly designed instructions to execute both steps simultaneously with appropriate few-shot demonstrations for inferring pseudo-interpretations. However, this approach performed significantly worse than our current method of separating the steps. The primary reason for this performance drop is that handling both tasks simultaneously imposes a substantial burden on a single LLM. Since each step requires detailed task descriptions and specific few-shot demonstrations, merging them results in an overload of information that negatively affects the model's reasoning process. This also highlights the challenges in effectively exploring various interpretations without an iterative approach (i.e., iterative RAG). Despite this, our proposed Retrieval Diversification (RD) method still efficiently infers pseudo-interpretations within the two-step framework and achieves strong results without the need for iterative processes. This emphasizes the effectiveness and efficiency of our reasoning chain design, even when tasks are separated for clarity and precision.

\subsection{Comparison to Verification Methods}
\label{ap:comp-verfication}

\subsubsection{Comparison to Chain-of-Verification} 
The verification modules in \proposed~and Chain-of-Verification (CoVe) \cite{dhuliawala2023chain} differ significantly in both their purpose ("why to use"), target (“where to use”), and timing ("when to use").

\textbf{Purpose ("why to use"):} The CoVe approach focuses on assessing the correctness of a generated response, ensuring the final output is accurate. In contrast, \proposed's verification module is designed to assess the relevance of retrieved passages within the RAG framework before any response is generated. These methods are therefore tailored for entirely different objectives—CoVe targets post-response accuracy, whereas \proposed~emphasizes pre-response relevance.

\textbf{Target (“where to use”):} The verifier in CoVe operates on the generated responses, whereas \proposed's verifier focuses on the retrieved passages. Given that these targets possess distinct characteristics and objectives, each verifier is uniquely designed to effectively capture the verification rationale relevant to its specific target. This fundamental difference between CoVe and \proposed~distinguishes the two approaches, making it challenging for the verifiers to be compatible or interchangeable.

\textbf{Timing ("when to use"):} CoVe performs verification after the response has been generated and presented to the user. In contrast, \proposed~operates earlier in the pipeline by verifying the retrieved passages before generating a response. Therefore, \proposed~is cost-efficient as it can anticipate whether a generated response is likely to be incorrect before the response is even produced. This allows \proposed~to avoid unnecessary response generation and associated costs, enabling the application of an optimal response strategy for the situation.

Additionally, \proposed~introduces a novel aspect in its verification, specifically designed for handling ambiguous queries. It employs pseudo-interpretations to evaluate how well the retrieved passages encompass multiple interpretations of the question. This approach is distinct from CoVe and further enhances the novelty of \proposed.

\subsubsection{Comparison to Verify-and-Edit} 
There are significant differences in the purposes of the verification modules within \proposed~and Verify-and-Edit \cite{zhao2023verify}. The Verify-and-Edit framework aims to assess the correctness of a generated chain of thought (CoT) and edit the CoT using retrieved external knowledge, ensuring an accurate reasoning process. On the other hand, \proposed’s verification module is tailored to evaluate the relevance of retrieved passages within the RAG framework.

These distinct goals highlight that the two methods are not only conceptually different but also serve different objectives. Therefore, our proposed verification module could be integrated into the Verify-and-Edit pipeline to improve the robustness of its editing process. Specifically, since the effectiveness of the Verify-and-Edit framework heavily relies on external knowledge for accurate editing, ensuring the relevance of retrieved passages is crucial. When there is ambiguity in the premise of the CoT, \proposed's verification module can verify the retrieved passages' relevance, enhancing the overall editing process.

\section{Additional Experiments}
\label{sec:ap:add-exp}

\subsection{Failure Cases}
\label{sec:ap:failure-cases}
We examined 50 randomly selected samples in ASQA from those with a D-F1 value below 0.5, where D-F1 ranges from 0 to 1.

The first failure scenario occurs when the retrieval diversification (RD) module underperforms, accounting for 22 out of the 50 samples. These failures arise from errors in the generated pseudo-interpretations or the inherent limitations of the base retriever (a frozen Sentence-BERT encoder). Utilizing fine-tuned retrievers, such as DPR, could partially alleviate this issue.

The second failure scenario occurs when the retrieval verification (RV) module underperforms. Ideally, the RV module should prioritize the LLM's internal knowledge over retrieved knowledge when the RD module underperforms. Hence, in this scenario, we identify cases where the RV incorrectly chooses retrieved knowledge instead of the LLM's internal knowledge under RD underperformance, resulting in errors. This accounts for 8 out of 22 samples, highlighting that our proposed RV module is not yet robust enough and leaves room for future research on developing a more accurate RV module for ambiguous questions. The remaining 14 out of 22 samples are particularly challenging and difficult for both RAG and closed-book LLMs.

The third failure scenario occurs when the RD module performs well, but the LLM fails to generate sufficiently accurate responses based on the retrieved passages. This accounts for 16 out of 50 samples and highlights an inherent limitation of the LLM rather than our framework, DIVA. Notably, there are no cases where the RV module underperforms when the RD module performs well.

The final failure scenario arises from other factors, such as the limitations of the evaluation metric D-F1, accounting for 12 out of 50 samples.

\subsection{Experiments on Unambiguous Questions}
\label{sec:unambiguous-experiments}
To assess the performance of \proposed~on unambiguous questions, we randomly selected 100 unambiguous questions from the NQ dataset. Specifically, we used the AmbigNQ dataset \cite{min2020ambigqa} to identify whether each question was ambiguous or unambiguous. Utilizing the identifier provided in the AmbigNQ dataset, we first isolated all unambiguous questions and then randomly sampled 100 questions from this subset for our evaluation. In Table~\ref{tab:qa_unambiguous}, our results indicate that the closed-book LLM achieves an EM score of 75, while incorporating the Vanilla RAG framework boosts the EM score to 80. Significantly, \proposed~outperforms both the closed-book LLM and Vanilla RAG, achieving QA performance on par with Iterative RAG. As highlighted throughout the paper, \proposed~is also twice as efficient as Iterative RAG while delivering comparable performance. These results confirm that \proposed, while tailored for ambiguous queries, also demonstrates strong performance on unambiguous ones, showcasing its broad applicability.

\begin{table}[ht]
    \centering
    \resizebox{0.5\columnwidth}{!}{\begin{tabular}{c|c}
        \toprule
         \textbf{Method} & \textbf{EM} \\
        \midrule
         Closed-book LLM & 75.0 \\
         Vanilla RAG & 80.0 \\
        Iterative RAG & \textbf{83.0} \\
        \midrule
        \proposed & \textbf{83.0} \\
        
        \bottomrule        
    \end{tabular}}
    \caption{QA performance on unambiguous questions. }
    \label{tab:qa_unambiguous}
\end{table}

Additionally, we assessed the effectiveness of \proposed’s Retrieval Diversification (RD) module on unambiguous queries using Recall@5. In Table~\ref{tab:ap-retrieval}, Vanilla RAG (row 1) refers to the baseline approach, where passages are retrieved based on a given question. + RD (Ours) applies our RD method to the baseline, incorporating pseudo-interpretations generated by \proposed~to diversify retrieval.
The results indicate that the RD module does not hinder retrieval performance. In fact, as shown in the table, it significantly improves Recall@5 when compared to Vanilla RAG. These findings demonstrate that RD effectively addresses issues of low-quality retrieval, enhancing performance for both ambiguous and unambiguous queries.

\begin{table}[ht]
    \centering
    \resizebox{0.6\columnwidth}{!}{\begin{tabular}{c|c|c}
        \toprule
        \textbf{Row} & \textbf{Method} & \textbf{Recall@5} \\
        \midrule
         1 & Vanilla RAG & 84.0 \\
        2 & + RD (Ours) & \textbf{87.0} \\
        
        \bottomrule        
    \end{tabular}}
    \caption{Retrieval accuracy on unambiguous questions. }
    \label{tab:ap-retrieval}
\end{table}

\subsection{Statistical Significance Test}

To verify that \proposed~consistently outperforms Vanilla RAG, we conduct a statistical significance test on the D-F1 metric. Given the cost of GPT API calls, we use GPT-3.5-turbo as the backbone model. A t-test is performed based on five experimental runs. The average D-F1 scores for Vanilla RAG and \proposed~are 36.8 and 38.1, respectively. The resulting p-value of the t-statistic is 0.0242, which is significantly below the 0.05 threshold, confirming that \proposed~achieves statistically significant improvements over Vanilla RAG. 

Due to cost constraints, we were limited to five runs. However, we observed that each additional run led to a gradual decrease in the p-value. This suggests that with more runs, the p-value would likely decrease further, providing even stronger statistical evidence for \proposed's superiority.

\section{Prompts}
\label{sec:ap:prompt}

Table~\ref{tab:ap:prompt_type_reason_infer} and Table~\ref{tab:ap:prompt_pseudo_inter} show an example of text prompt for inferring \textit{pseudo-interpretations} (i.e., $I_{\text{a}}$ and $I_{\text{p}}$ in Eqn~\ref{eq:infer-pseudo-interpretations}). Table~\ref{tab:ap:retrievel_verify} shows an example of text prompt for verifying the retrieved passages (i.e., $I_{\text{v}}$ in Eqn~\ref{eq:dq_verify}). Table~\ref{tab:ap:prompt_extract_interpretations_from_passage} and \ref{tab:ap:prompt_rag_response_generation} show an exmple of text prompt for response generation in vanilla RAG framework (i.e., $I_{\text{e}}$ in Eqn~\ref{eq:disambiguation} and $I_{\text{g}}$ in Eqn~\ref{eq:question-answering})

\begin{table*}[h]
\captionof{table}{Example of Prompt $I_{\text{a}}$.}
\vspace{-1ex}
\small
\centering
\begin{tabular}{|l|}
\hline
\rowcolor{grey}\multicolumn{1}{|c|}{\textbf{Instruction}}   \\

Your task is to determine which types of ambiguity are related to a given question. Types of ambiguity in a \\ question can be defined as follows: \\
\\
1. [AmbSub]: This type of ambiguity arises when the subject of the question is not clear. The subject is the \\ person, place, thing, or idea that is doing or being something. It's the entity about which information is \\ being sought. \\
2. [AmbObj]: This type of ambiguity arises when the object of the question is unclear. The object\\ refers to the entity that the action or state expressed by the verb is directed towards. \\
3. [AmbPred]: This type of ambiguity arises when the predicate of the question is unclear. The predicate is \\ the part of a sentence that tells us what the subject does or is. It includes the verb and everything else  \\ that comes after the subject. \\
4. [AmbTime]: This type of ambiguity arises when the time frame of the question is unclear. This can lead to \\ confusion because many actions or states can change over time. \\
5. [AmbLoc]: This type of ambiguity arises when the location referred to in the question is unclear. Many \\ events or entities can exist in different locations, leading to confusion. \\
6. [N/A]: This type of ambiguity arises when there is no ambiguous point in the given question. \\
\\
Below are some examples that map the question to the types.\\
\\
Question: Who has scored the most goals in international soccer\\
Types: [AmbSub]. The subject "Who" may refer to either men or women.\\
\\
Question: What is the date of the queen's birthday?\\
Types: [AmbObj]. The object "the date of the queen's birthday" may refer to the date of Queen Elizabeth II's \\birthday or Queen Victoria's birthday.\\
\\
Question: Who appeared in the Wimbledon finals 2017?\\
Types: [AmbPred]. The predicate "appeared" could refer to the tennis players or celebrities in the audience.\\
\\
Question: Where is the u21 euro championships being held?\\
Types: [AmbTime]. You may need to clarify whether it refers to the championships being held in 2015, 2017, \\ or 2019.\\
\\
Question: When is the new iPhone being released?\\
Types: [AmbLoc]. This may need clarification on whether it refers to the release date in the United States, \\Europe, Asia, or another region.\\


\rowcolor{grey}\multicolumn{1}{|c|}{\textbf{Few-shot Demos}}   \\

Given the ambiguous question that can be interpreted in multiple ways, which types of ambiguity are related \\ to the question? Suggest the types and provide reasons for your suggestions. Please use the format of: \\ \#\#Reason: \{reason\} \#\#Answer: \{answer\}.\\
\\
question: Who is top goalscorer in the world cup?\\
\\
\#\#Reason: The subject "Who" in the question may refer to either men or women, as both men's and women's \\ FIFA World Cups are held. \\
\#\#Answer: [AmbSub]\\

\rowcolor{grey}\multicolumn{1}{|c|}{\textbf{Actual Question}}   \\
\\
Given the ambiguous question that can be interpreted in multiple ways, which types of ambiguity are related \\ to the question? Suggest the types and provide reasons for your suggestions. Please use the format of: \\\#\#Reason: \{reason\} \#\#Answer: \{answer\}.\\
\\
question: Who has the highest goals in world football?\\
\texttt{\#\#Reason: The subject \"Who\" in the question is ambiguous as it may refer to either men or} \\ \texttt{women.} \\ \texttt{The disambiguation clarifies this by specifying the gender.} \\ \texttt{\#\#Answer: [AmbSub]} \\

\bottomrule

\end{tabular}
\label{tab:ap:prompt_type_reason_infer}
\end{table*}

\begin{table*}[h]
\captionof{table}{Example of Prompt $I_{\text{p}}$.}
\vspace{-1ex}
\centering
\begin{tabular}{|l|}
\hline
\rowcolor{grey}\multicolumn{1}{|c|}{\textbf{Instruction}}   \\

I will provide an ambiguous question that can have multiple answers based on different possible \\interpretations. Additionally, I will provide corresponding reasons why the question is ambiguous. \\ Clarify the given question into several disambiguated questions based on the reasons for its ambiguity. \\ Please use the format of: \#\#Disambiguations: \{disambiguations\}: \\

\rowcolor{grey}\multicolumn{1}{|c|}{\textbf{Few-shot Demos}}   \\

\#\#Question: Who is top goalscorer in the world cup?\\
\\
\#\#Reason: The subject "Who" in the question may refer to either men or women, as both men's and\\  women's FIFA World Cups are held.\\
\\
\#\#Disambiguations:\\
1: Who is the top goalscorer in the men's FIFA world cup?\\
2: Who is the top goalscorer in the women's FIFA world cup?\\

\rowcolor{grey}\multicolumn{1}{|c|}{\textbf{Actual Question}}   \\
\\
\#\#Question: Who has the highest goals in world football?\\
\\
\#\#Reason: The subject "Who" in the question is ambiguous as it may refer to either men or women. \\ The disambiguation clarifies this by specifying the gender.\\\\
\#\#Disambiguations: \\
\texttt{1: Which male player has the highest goals in world football?} \\ \texttt{2: Which female player has the highest goals in world football?} \\

\bottomrule

\end{tabular}
\label{tab:ap:prompt_pseudo_inter}
\end{table*}

\begin{table*}[h]
\captionof{table}{Example of Prompt $I_{\text{v}}$.}
\vspace{-1ex}
\centering
\begin{tabular}{|l|}
\hline
\rowcolor{grey}\multicolumn{1}{|c|}{\textbf{Instruction}}   \\

Given the question and its relevant passages,  determine whether the passage contains the answer to the \\ question. Please answer with Yes or No. \\

\rowcolor{grey}\multicolumn{1}{|c|}{\textbf{Actual Question}}   \\
\\
Question: Which male player has the highest goals in world football?'
\\
Passage: \\
$[1]$ List of footballers with the most goals in a single game | This is a list of players with the most goals \\ in a football game...\\
...\\
$[5]$ List of men's footballers with 50 or more international goals | In total, 79 male footballers to date \\ have scored at least 50 goals with their national team at senior level ... \\

\\
Response:\\
\texttt{Yes.}
\\

\bottomrule

\end{tabular}
\label{tab:ap:retrievel_verify}
\end{table*}

\begin{table*}[h]
\captionof{table}{Example of Prompt $I_{\text{e}}$.}
\vspace{-1ex}
\centering
\begin{tabular}{|l|}
\hline
\rowcolor{grey}\multicolumn{1}{|c|}{\textbf{Instruction}}   \\\\
I will provide ambiguous questions that can have multiple answers based on their different possible \\ interpretations. Clarify the given question into disambiguated questions as many as possible and \\ provide short factoid answers to each question. Subsequently, summarize them into a detailed \\ long-form answer of at least three sentences. Here are some examples.\\

\\

\rowcolor{grey}\multicolumn{1}{|c|}{\textbf{Few shot demos}}   \\
\\
Question: Who is top goalscorer in the world cup?\\
\\
Disambiguations:\\
DQ 1: Who is the top goalscorer in the men's FIFA world cup?\\
DA 1: Miroslav Klose\\
DQ 2: Who is the top goalscorer in the women's FIFA world cup?\\
DA 2: Marta\\
\\
Answer: There is a men's and women's FIFA world cup tournament. The the top goalscorer in the \\ men's FIFA world cup is German player, Miroslav Klose. And in the women's division, the top  \\ goalscorer is a Brazilian player, most commonly known as Marta.
\\\\

\rowcolor{grey}\multicolumn{1}{|c|}{\textbf{Actual Question}}   \\
\\
Context:\\
$[1]$ List of footballers with 500 or more goals | Bican (Slavia Prague), Jimmy Jones (Glenavon), ... \\ 
...\\
$[5]$ List of men's footballers with 50 or more international goals | In total, 79 male footballers ... \\ 
\\
Question: Who has the highest goals in world football? \\
\\
Disambiguations:\\
\texttt{DQ 1: Who has scored the highest number of goals in their football career in} \\ \texttt{history?} \\
\texttt{DA 1: Josef Bican}\\
\texttt{DQ 2: Who has scored the highest number of goals in a single football game?}\\
\texttt{DA 2: Archie Thompson; Shokhan Nooraldin Salihi}\\
\texttt{DQ 3: Who has scored the highest number of international goals in men's football?}\\
\texttt{DA 3: Cristiano Ronaldo}\\
\texttt{DQ 4: Who has scored the highest number of goals in a calendar year in} \\ \texttt{international matches?}\\
\texttt{DA 4: Cristiano Ronaldo}\\

\bottomrule

\end{tabular}
\label{tab:ap:prompt_extract_interpretations_from_passage}
\end{table*}

\begin{table*}[h]
\captionof{table}{Example of Prompt $I_{\text{g}}$}
\vspace{-1ex}
\centering
\begin{tabular}{|l|}
\hline
\rowcolor{grey}\multicolumn{1}{|c|}{\textbf{Instruction}}   \\
I will provide ambiguous questions that can have multiple answers based on their different possible \\ interpretations. Clarify the given question into disambiguated questions as many as possible and \\ provide short factoid answers to each question. Subsequently, summarize them into a detailed \\ long-form answer of at least three sentences. Here are some examples. \\
\\

\rowcolor{grey}\multicolumn{1}{|c|}{\textbf{Few shot demos}}   \\
\\
Question: Who is top goalscorer in the world cup?\\
\\
Disambiguations:\\
DQ 1: Who is the top goalscorer in the men's FIFA world cup?\\
DA 1: Miroslav Klose\\
DQ 2: Who is the top goalscorer in the women's FIFA world cup?\\
DA 2: Marta\\
\\
Answer: There is a men's and women's FIFA world cup tournament. The the top goalscorer in the \\ men's FIFA world cup is German player, Miroslav Klose. And in the women's division, the top \\ goalscorer is a Brazilian player, most commonly known as Marta.
\\

\rowcolor{grey}\multicolumn{1}{|c|}{\textbf{Actual Question}}   \\
\\
Context:\\
$[1]$ List of footballers with 500 or more goals | Bican (Slavia Prague), Jimmy Jones (Glenavon), ... \\ 
...\\
$[5]$ List of men's footballers with 50 or more international goals | In total, 79 male footballers ... \\ 
\\
Question: Who has the highest goals in world football?
\\
Disambiguations:\\
DQ 1: Who has scored the highest number of goals in their football career in history? \\
DA 1: Josef Bican\\
DQ 2: Who has scored the highest number of goals in a single football game?\\
DA 2: Archie Thompson; Shokhan Nooraldin Salihi\\
DQ 3: Who has scored the highest number of international goals in men's football?\\
DA 3: Cristiano Ronaldo\\
DQ 4: Who has scored the highest number of goals in a calendar year in international matches?\\
DA 4: Cristiano Ronaldo\\
\\
Answer:\\
\texttt{The question "Who has the highest goals in world football?" can be interpreted} \\  
\texttt{in several ways. If we consider the highest number of goals scored in ...}  \\

\bottomrule

\end{tabular}
\label{tab:ap:prompt_rag_response_generation}
\end{table*}

\begin{table*}[h]
\captionof{table}{Example of Prompt $I_{\text{l}}$}
\vspace{-1ex}
\centering
\begin{tabular}{|l|}
\hline
\rowcolor{grey}\multicolumn{1}{|c|}{\textbf{Instruction}}   \\
I will provide ambiguous questions that have multiple answers regarding different aspects of the \\ question. Your task is to generate an answer that includes as many aspects as possible from the \\ ambiguous questions. \\

\rowcolor{grey}\multicolumn{1}{|c|}{\textbf{Few shot demos}}   \\
\\
Question: Who is top goalscorer in the world cup?\\
\\
Given the question, generate a comprehensive long-form answer.\\
\\
Final Answer: There is a men's and women's FIFA world cup tournament. The the top goalscorer \\ in the men's FIFA world cup is German player, Miroslav Klose. And in the women's division, the \\ top goalscorer is a Brazilian player, most commonly known as Marta. \\

\rowcolor{grey}\multicolumn{1}{|c|}{\textbf{Actual Question}}   \\
\\
Question: Who has the highest goals in world football?\\
\\
Given the question, generate a comprehensive long-form answer.\\
\\
Final Answer:\\
\texttt{The highest goals in world football can be interpreted in different ways. If we} \\ 
\texttt{are talking about the highest number of goals scored in a professional football} \\ 
\texttt{career, the record belongs to Josef Bican, who scored an estimated 805 goals in} \\ \texttt{competitive ...} \\
\bottomrule

\end{tabular}
\label{tab:ap:prompt_closedbookllm_response_generation}
\end{table*}

\end{document}